\documentclass[]{interact}

\usepackage[natbibapa,nodoi]{apacite}%
\usepackage[T1]{fontenc}
\usepackage{amsmath,amsfonts} %
\usepackage{algorithm}
\usepackage{algpseudocode}
\usepackage{textcomp}
\usepackage{xcolor}
\usepackage{comment}
\usepackage{hyperref}
\hypersetup{
    pdfauthor={Sohaib}, %
	colorlinks=true,
	linkcolor=magenta,
	citecolor=blue,
	filecolor=green,
	urlcolor=cyan,
}
\usepackage{booktabs}
\usepackage{parskip}
\usepackage{graphicx} %
\usepackage{graphics}
\usepackage{caption}
\usepackage{subcaption}
\usepackage[normalem]{ulem} %
\usepackage{soul} %
\usepackage{mathtools}			%
\usepackage{bm,nicefrac,xfrac}
\usepackage{relsize} 
\usepackage{array} 
\usepackage{bbm}
\usepackage{setspace}
\usepackage{multirow}
\usepackage{siunitx}
\usepackage[bottom]{footmisc}
\usepackage{tikz}
\usetikzlibrary{positioning}
\usepackage{pgfplots}
\pgfplotsset{compat=newest}
\usepackage{microtype}

\usepackage{lineno}
\modulolinenumbers[1]

\newtheorem{remark}{Remark}

\newcommand*{\TwoRows}[2]{{\begin{tabular}[c]{@{}c@{}} {#1} \\ {#2} \end{tabular}}}

\begin{document}

\title{CAMP: A Context-Aware Cricket Players Performance Metric} %

\author{
\name{Muhammad Sohaib Ayub\textsuperscript{a}, Naimat Ullah\textsuperscript{a}, Sarwan Ali\textsuperscript{b}, Imdad Ullah Khan\textsuperscript{a}*\thanks{* Corresponding author}, \\Mian Muhammad Awais\textsuperscript{a}, Muhammad Asad Khan\textsuperscript{c} and Safiullah Faizullah\textsuperscript{d}}
\affil{\textsuperscript{a}Department of Computer Science, LUMS, Pakistan\\
\textsuperscript{b}Department of Computer Science, Georgia State University, USA\\
\textsuperscript{c}Department of Telecommunication, Hazara University, Pakistan\\
\textsuperscript{d}Department of Computer Science, Islamic University, KSA}
\thanks{E-mail addresses: 15030039@lums.edu.pk (MS. Ayub), 18030048@lums.edu.pk (N. Ullah), sali85@student.gsu.edu (S. Ali), imdad.khan@lums.edu.pk (IU. Khan), awais@lums.edu.pk (MM. Awais), asadkhan@hu.edu.pk (MA. Khan), safi@iu.edu.sa (S. Faizullah)}
}
\maketitle

\begin{abstract}
Cricket is the second most popular sport after soccer in terms of viewership. However, the assessment of individual player performance, a fundamental task in team sports, is currently primarily based on aggregate performance statistics, including average runs and wickets taken. We propose \textbf{C}ontext-\textbf{A}ware \textbf{M}etric of player \textbf{P}erformance, \textsc{camp}, to quantify individual players' contributions toward a cricket match outcome. \textsc{camp} employs data mining methods and enables effective data-driven decision-making for selection and drafting, coaching and training, team line-ups, and strategy development. \textsc{camp} incorporates the exact context of performance, such as opponents' strengths and specific circumstances of games, such as pressure situations. We empirically evaluate \textsc{camp} on data of limited-over cricket matches between 2001 and 2019. In every match, a committee of experts declares one player as the best player, called {\em \textbf{M}an \textbf{o}f the \textbf{M}atch} (\textsc{MoM}). The top two rated players by \textsc{camp} match with \textsc{MoM} in 83\% of the 961 games. Thus, the \textsc{camp} rating of the best player closely matches that of the domain experts. By this measure, \textsc{camp} significantly outperforms the current best-known players' contribution measure based on the Duckworth-Lewis-Stern (\textsc{dls}) method.

\end{abstract}


\section{Introduction}
Analysis of fine-grained sports data plays a pivotal role in data-driven decision-making in all aspects of sports management~\citep{fried2016sport}. Many machine learning models have been proposed for game modeling and match outcome prediction for soccer~\citep{Tom2018Actions,davis2019assessing,lu2022what}, basketball~\citep{Sameer2016NBAplayers}, and hockey~\citep{lord2022field,Guiliang2018IceHockey}. However, data-driven decision-making has not received much attention in cricket, which has the second-highest viewership~\citep{Vignesh14} after soccer and is a multi-billion dollar industry.

In addition to tournaments organized by the International Cricket Council (ICC)\footnote{We provide a brief overview of the cricket game with the terminology and rules of the game in Appendix~\ref{appendix_section}. Detailed information regarding cricket is also available online \url{https://www.icc-cricket.com/about/cricket/rules-and-regulations/playing-conditions}}, numerous cricket leagues and regional and inter-departmental games are played across the globe. A fundamental task at every level and game aspect is to measure players' quality and worth. All the key stakeholders of the game (e.g., selectors, coaches, franchise owners, and even brand managers) are often interested in the following question: \textit{How much does the performance of an individual player impact the outcome of a given match~\citep{Tom2018Actions}?} Players' performance assessment helps franchise owners and selectors in drafting contracts, sports bodies in talent hunt, coaches to determine optimal bowler versus batter matchups, and brand managers to organize media promotions.

Currently, performance assessment in cricket is primarily made by experts based on qualitative judgments by scrutinizing the entire match situation. These judgments rely on aggregate statistics of standard performance measures. However, these measures of batting and bowling performance (e.g., batting average, batting strike rate~\citep{barr2004criterion}, bowling economy\footnote{The batting strike rate is the percentage of runs scored from the balls faced by the batter ($\nicefrac{runs}{balls} * 100$). The bowling economy is the number of runs conceded by the bowler per over ($\nicefrac{runs}{overs}$).}) have three significant limitations. Firstly, these measures assign a fixed value to each achievement~\citep{davis2015player,Stern09}, regardless of the specific opponent against whom the achievement was made. For instance, for bowlers, wickets are considered equivalent irrespective of the batters' quality, and for batters, runs scored carry equal weight regardless of the bowlers' strength. Secondly, these measures do not account for the stage of the innings, such as pressure index~\citep{shah2014pressure,bhattacharjee2016quantifying,bhattacharjee2020predicting}. Lastly, they only consider immediate effects and do not incorporate the downstream impact. For example, the early wicket loss of an opening batter also reduces the team's overall capability to score runs.

Data analysis on fine-grained cricket data can highlight slim differences in skills and performance imperceptible to a human. Actionable analytics drawn from data will aid `managers' in optimal decision-making, reduce players' contract costs, increase efficiency, and minimize bias. Some data analytics work has been done to quantify players' performance~\citep{Lewis05,lewis2008extending} and a pair of batters~\citep{LemmerBattingMeasure2018}. However, these approaches only consider the remaining resources (remaining overs and wickets) as game context, whereas qualitative aspects of remaining players and resources also contribute to important contextual information. 

In this paper, we propose a novel tool, \textbf{C}ontext \textbf{A}ware \textbf{M}etric of player \textbf{P}erformance (\textsc{camp}), to rate the players by measuring their contributions considering the context of the game. Unlike the current state of the art work, referred to as \textbf{L}ewis \textbf{N}et \textbf{C}ontribution (\textsc{lnc})~\citep{Lewis05}, we also consider additional features like the quality of the remaining resources and performance made so far by a team as the game context. \textsc{camp} calculates each player's contribution score incorporating the game venue, the stage of the match, the opposing players, and the overall strength of the opposition team. 

We estimate the expected runs to be scored by the batting team at every stage of the game, using a combination of supervised and unsupervised machine learning techniques. We use current match information and historical game data to capture context about \textit{similar} performing teams and players. Based on the expected and actual runs scored in an over, we measure over-by-over players' contribution, which is aggregated for players' ratings at the match level. 

We compare \textsc{camp} players' ratings with the ICC announced \textbf{M}an \textbf{o}f the \textbf{M}atch (referred as \textsc{MoM}) and \textsc{lnc}~\citep{Lewis05}. We show that the experts' opinion-based top-rated player (\textsc{MoM}) substantially agrees with one of the \textsc{camp}'s top-rated players. This indicates that at least at one end of the spectrum, \textsc{camp} successfully emulates domain experts. 
While our approach can be used for any format of the game, in this paper, we focus on one of the limited-over formats known as \textit{One Day International} (\textsc{odi}). 

The main features of this work are the following:
\begin{itemize}

\item We propose \textsc{camp} that quantify the contributions of all $22$ players in a cricket match. It computes rating considering the context of the match (opposition strength, stage of the innings). Various stakeholders (selectors, coaches, franchise owners, brand managers) can use \textsc{camp} for effective decision-making.
\item As a subroutine, we develop a model that predicts projected runs at any stage of the game (i.e., runs the batting team can score in the remaining part of the game). This model is helpful for strategy adjustments during a live game, legal betting applications and may be of independent research interest. 

\item The results show that the performance score by \textsc{camp} agrees with that of experts' decision of \textsc{MoM} to a greater extent as \textsc{MoM} is the top-rated player and one of the top two rated players by \textsc{camp} in $66\%$ and $83\%$ of the games, respectively. \textsc{camp} also outperforms the state of the art approach \textsc{lnc} based on the Duckworth-Lewis-Stern (\textsc{dls}) method. 

\item \textsc{camp} ratings at match level can be extended to series level (a set of consecutive matches) and career level to estimate the \textit{net worth} of a player. These estimates are of particular interest to international cricket bodies and franchise owners. 

\item We perform experiments on a comprehensive dataset of $961$ \textsc{odi} matches played between $2001$ and $2019$. We make the preprocessed dataset publicly available, opening up a broad avenue of further research in cricket data analytics.

\end{itemize}

The rest of the paper is organized as follows. Section~\ref{related_work} briefly reviews the literature on sports data analytics. Section~\ref{proposedApproach} presents our proposed approach \textsc{camp}. We give the detailed experimental setup in Section~\ref{experimentalSetup}. We present the empirical results in Section~\ref{resultsAndDiscussion} and conclude the paper in Section~\ref{conclusion}.

\section{Related Work}\label{related_work}
Quantifying the impact of players' performance is a well-studied problem in sports data analysis, particularly for basketball~\citep{Sameer2016NBAplayers}, soccer~\citep{Tom2018Actions,lu2022what}, and hockey~\citep{Guiliang2018IceHockey}. 

Several machine learning models have been proposed for game modeling and outcome prediction, ranging from simple supervised and unsupervised learning to graphical models~\citep{Adrian2006Football,Rory2017ML}. Dolores adopts a neural network-based approach using dynamic ratings and Bayesian networks for predicting the outcome of football matches~\citep{Constantinou2019}. Outcome prediction in sports is generally treated as a classification problem with two or three classes (win, lose, or draw)~\citep{Prasetio2016PredictingFM,Zimmermann2013PredictingNM,bhattacharjee2020predicting}. However, few studies have used regression-based approaches to predict game outcome~\citep{Delen2012ACA,Goddard2005}. These studies also predict victory margins (e.g., the difference between the number of goals scored by each team in a soccer game). 

Although many popular sports are well-studied in the literature, cricket remains unexplored mainly due to the game's dynamic and unpredictable nature. The Duckworth-Lewis (\textsc{dl}) method~\citep{Duckworth1998} is a technique to reset the batting targets for interrupted limited-overs matches. Adopted in $1999$ by ICC as the official target resetting method, \textsc{dl} method is based on a resource table where each entry represents the percentage of resources available to the batting team. The main limitation of the \textsc{dl} method is using the same resource table for both innings, whereas scoring patterns in the second innings differ significantly from the first.
Factors such as the pressure of chasing contribute to the fact that the first innings cannot be directly compared to the second innings. To overcome this problem,~\citep{Stern09} extended the \textsc{dl} method, known as the Duckworth-Lewis-Stern (\textsc{dls}) method, and proposed a separate resource table for the second innings. 

\cite{clarke1988dynamic} used dynamic programming to model cricket game progression. For any stage of the first innings, he proposes a dynamic programming-based optimal scoring rate along with an estimated total number of runs that would be scored. For each stage of the second innings, he models the probability of winning considering wickets in hand, number of overs remaining, and runs yet to be scored.~\citep{Beaudoin03} developed a new technique for analyzing team performance and finding the most valuable players using the \textsc{dl} resource table.~\citep{Lemmer08} proposes an approach that assigns weights to traditional performance measures (such as batting averages, count of scores while remaining not-out, and bowling averages) to analyze the players' performance.~\citep{Jhawar16} uses various features of batters and bowlers to predict the match outcome using the nearest neighbor classifier. The pressure index~\citep{bhattacharjee2016quantifying,shah2014pressure,bhattacharjee2020predicting} captures the changing circumstances of matches for measuring the performance of cricketers. The study conducted by~\citep{saikia2019performing} demonstrates the potential of pressure index for measuring performance, comparing run chases and predicting match outcomes.~\citep{Lewis05} proposed \textsc{lnc} to measure player performance using the \textsc{dl} resource table. Based on the percentage of the resources remaining at any stage of an inning, \textsc{lnc} estimates the expected runs to be scored. Players' contribution is then estimated from expected runs and actual runs scored. This approach relies on the \textsc{dl} resource table, which is too general and does not consider the match-specific details.

Various works incorporate historical information to predict match outcomes and suggest suitable team combinations. 
A combination of linear regression and the nearest neighbor algorithm predicts the winning team by estimating the runs to be scored in the innings' remaining part, and the estimated runs are updated based on historical and current match data after an interval of $5$ overs~\citep{Vignesh14}. An approach suggests a suitable team combination by applying association rule mining on historical players' performance~\citep{Bhattacherjee15,norman2010optimal,Tim2006OptimalBattingOrder}. The power play also significantly impacts the match outcome, and teams perform better during power play overs in terms of run rate, wicket preservation, and ultimately, match win probability~\citep{bhattacharjee2016impact}.
Teams strength is analyzed based on the players' historical performance, and match outcome is predicted based on current match data in \textsc{t20} format~\citep{Viswanadha17,bhattacharjee2016impact}, \textsc{test}~\citep{scarf2011analysis} and \textsc{odi}~\citep{hasanika2021data} cricket. Similarly, the study by~\citep{saikia2019cricket} performed the statistical analysis of various performance measures to quantify the performance of cricketers including batters~\citep{lemmer2011single}, bowlers~\citep{saikia2012predicting}, fielders~\citep{bhattacharjee2022quantification} and all-rounders~\citep{saikia2011classification}. These  batting, bowling and fielding performance indicators can be used  to obtain a competitive balance in team selection~\citep{bhattacharjee2016objective,bhattacharjee2014performance}.

\section{Proposed Approach: The \textsc{camp} Algorithm}\label{proposedApproach}
\begin{figure}[h!]
    \centering
    \includegraphics[scale=0.66]{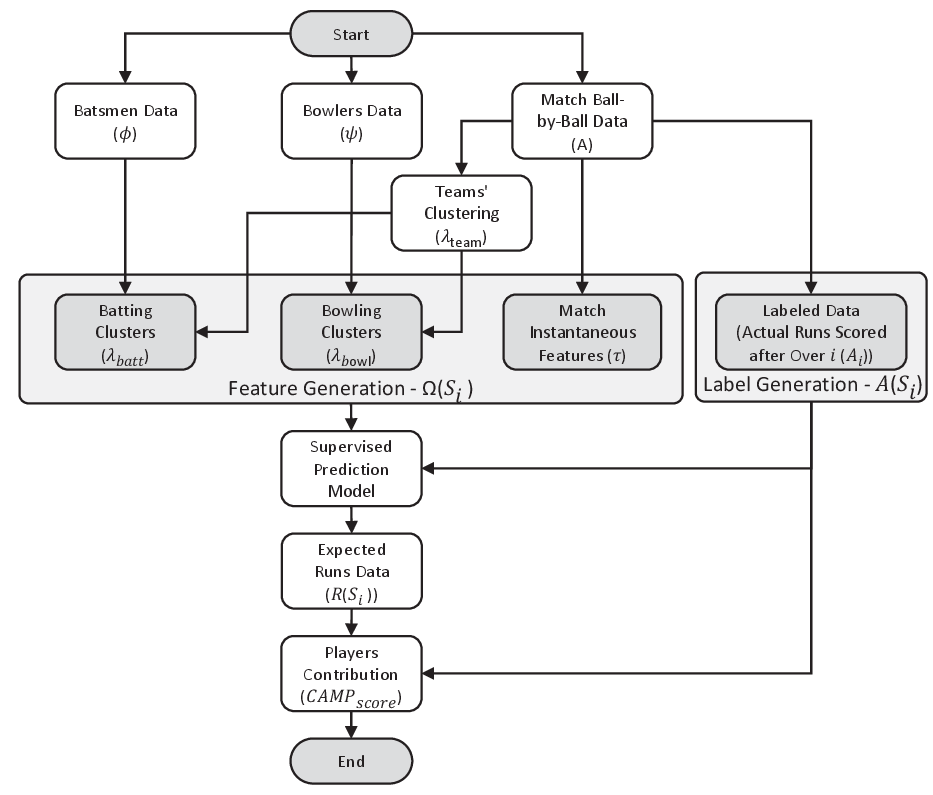}
    \caption{Flow diagram of our proposed model \textsc{camp}.} %
    \label{fig_Flow_Diagram}
\end{figure}

In this section, we formulate the problem of quantifying players' contributions from ball-by-ball \textsc{odi} matches data. We estimate the expected runs for each over using the current game's status, teams' strength, players' quality, and match venue. \textsc{camp} computes the contributions of the players (batters and bowlers) based on the difference between expected runs and actual runs scored. For simplicity, we divide our problem into the following two sub-problems: 

\begin{enumerate}
    \item Estimation of expected runs to be scored in any over at a given stage of the innings. The challenging part of this problem is to capture the context of the game, including the players' quality determined by players' past game history, teams' strength, match venue, and remaining resources. Moreover, It also requires avoiding the cold-start problem to capture players' quality.

Due to the cold-start problem, data sparsity hinders learning the players' features. 
A significant challenge in the accurate computation of the expected score is limited ({\em `data sparsity'}) or no available data ({\em `cold start'} problem). Given the amount and timeline of data, a given batter $b$ may have no or very sparse playing history against a bowler $l$~\citep{Vignesh14}. Thus, a machine learning model may not be able to learn any valuable insight for prediction. 
Therefore, we cluster the batters and bowlers to tackle the cold-start and data sparsity problem as similar batters or bowlers can be considered in place of a specific query batter or bowler. We empirically validate the players' clustering in Section~\ref{playerClusterValidation} and Section~\ref{playerstsne}.

    \item Computation of players' ratings based on the expected runs and actual runs scored in an over. The challenging part of this problem is finding players' ratings confirming the experts' decision-based top-rated player (\textsc{MoM}).
\end{enumerate}
A list of frequently used symbols with their description is given in Table~\ref{tbl_notations}. We provide an overview of \textsc{camp} in Figure~\ref{fig_Flow_Diagram}, and each step is explained in the following sections. 

\begin{table}[h!]
    \centering
     \begin{tabular}{ll}
        \toprule
        {Symbol} & {Description}\\
        \midrule
        \textbf{$S_i$} & {Innings stage at start of over $i$}\\[.015in]
        \textbf{$P(S_i)$} & {Projected total runs estimated at $S_i$}\\[.015in]
        \textbf{$T(S_i)$} & {Total runs scored till $S_i$}\\[.015in]
        \textbf{$R(S_i)$} & {Projected remaining runs at $S_i$. Runs to be scored after $S_i$}\\[.015in]
        \textbf{$A(S_i)$} & {Actual runs scored after $S_i$}\\[.015in]
        \textbf{$r_i$} & {Total runs scored in over $i$}\\[.015in]
        \textbf{$r_i^{p}$} & {Runs scored by player $p$ in over $i$}\\[.015in]
        \textbf{$e_i$} & {Expected runs in over $i$}\\[.015in]
        \textbf{$c_i^{p}$} & {Contribution by player $p$ in over $i$ }\\[.015in]
        \textbf{$C_{bat}(p)$} & {Aggregated batting contribution for player $p$ in a match}\\[.015in]
        \textbf{$C_{bowl}(p)$} & {Aggregated bowling contribution for player $p$ in a match}\\[.015in]
        \textsc{camp$_{score}$} & {Net contribution vector for all participating players}\\[.015in]
        \textbf{$\phi_p$} & {Batters feature vector for player $p$}\\[.015in]
        \textbf{$\psi_p$} & {Bowlers feature vector for player $p$}\\[.015in]
        \textbf{$\Omega(S_i)$} & {Feature vector to predict $R(S_i)$ at $S_i$ }\\
        \bottomrule
    \end{tabular}
    \caption{Notations used in our proposed model \textsc{camp}.}
    \label{tbl_notations}
\end{table}

\subsection{Projected Score Computation}
This section describes our methodology to compute projected remaining runs using historical data and current match information for both teams, including their participating players and venue. 
We define $S_i$ as the stage of innings at the start of over $i$, $1\le i \le 50$. The projected remaining runs at $S_i$ are represented by $R(S_i)$.
To capture the qualitative aspect of resources (overs, wickets) in $S_i$, we represent teams and players as feature vectors and cluster them into performance-based groups. These teams' and players' clusters, along with current match data, are used to generate match stage feature vectors $\Omega(S_i)$, which are used to predict $R(S_i)$. 

\subsubsection{Teams' Clustering} \label{teamsclustering}
\begin{figure}[h!]
    \centering
    \includegraphics[scale=0.7]{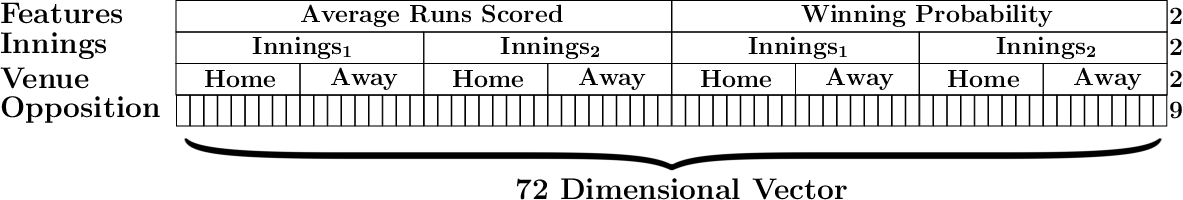}
    \caption{Teams feature vector generated using historical batting data. Historical data of a team is collected against each of the $9$ ICC top-ranked teams for different features such as venue, innings, average runs scored, and winning probability.}
    \label{fig:teamsFeatureVector}
\end{figure}

We group ten regular and ICC-ranked teams into different clusters. This categorization of teams helps avoid the data sparsity problem (a new player having no historical information against specific players of other teams) in players' clustering. For this purpose, we design a $72$-d vector/embedding (based on the batting performance of teams) that contains the average runs scored and the team's winning probability against each of the $9$ opponent teams while playing both innings for both types of venues ``home/away". The feature vector is shown in Figure~\ref{fig:teamsFeatureVector}. 
The feature embeddings are then used as input to the standard $k$-means clustering algorithm to cluster the teams (where $k=3$, decided using the standard validation set approach~\citep{validationSetApproach}). The teams in different clusters are shown in Table~\ref{tab_teamsCluster}. 

\begin{remark}
We also clustered teams' by their bowling records (Average Runs Scored, Winning Probability, and Wickets Taken), but the clusters remained the same.
\end{remark}

\begin{table}[h!]
\small
    \centering
    \begin{tabular}{lllll}
        \toprule
        {Cluster ID} & \multicolumn{4}{c}{{Teams}} \\
        \midrule
        Cluster 1 & Australia (AUS) & England (ENG) & South Africa (SA) & Sri Lanka (SL) \\
        Cluster 2 & India (IND) & Pakistan (PAK) & Bangladesh (BAN) & - \\
        Cluster 3 & West Indies (WI) & New Zealand (NZ) & Zimbabwe (ZIM) & - \\
        \bottomrule
    \end{tabular}
    \caption{Top $10$ ICC ranked teams grouped into $3$ clusters to avoid the cold-start problem by considering the similar teams' cluster in place of a specific query team.}
    \label{tab_teamsCluster}
\end{table}

These teams' clusters are used in the players' feature vectors to avoid data sparsity problem by considering the similar team's cluster in place of a specific query team. %

\subsubsection{Batters Clusters} \label{battingclusters}

\begin{figure}[h!]
    \centering
    \includegraphics[scale=0.6]{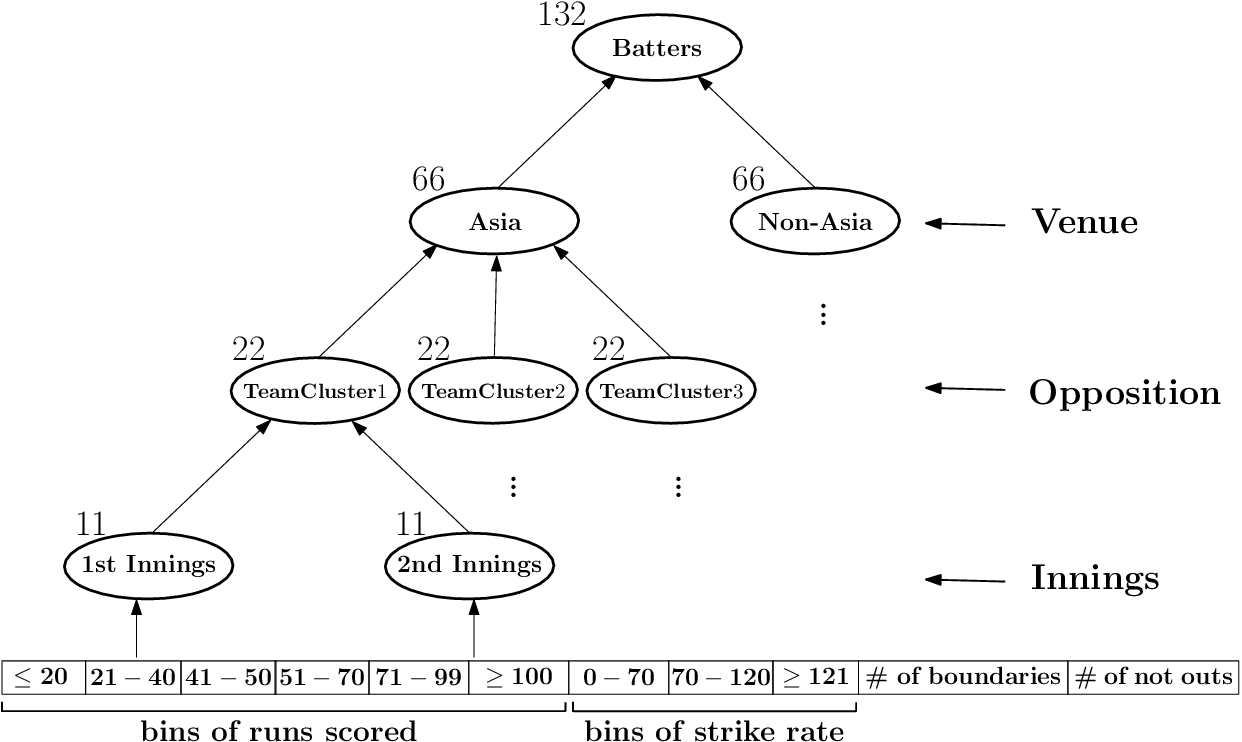}
    \vskip 0.1in
    \caption{A feature vector for a batter consists of $5$ bins for runs scored, $3$ bins for strike rate, count of boundaries, and count of not-outs. These $11$ performance parameters are recorded across different venues, oppositions, and innings to form a $132$-d feature vector.}
    \label{fig:batsman_feature_vector} %
\end{figure}

To cluster the players based on their batting quality, we represent each player by a feature vector comprised of past batting performances at different venues, against different oppositions (teams clusters) in the first or second innings (Figure~\ref{fig:batsman_feature_vector}). More formally, we form a feature embedding, $\phi_p$ for player $p$ based on the $11$ performance parameters. $\phi_p$ discretizes the runs scored and the strike rate into $6$ and $3$ bins, respectively, such that each bin contains the count of the corresponding value. We also record the total number of boundaries scored and the count of matches in which $p$ remains not-out. 

We keep these $11$ performance parameters for the granularity level of innings, opposition's strength and venue class. For the first granularity level, the match venue is categorized into two classes, Asia and non-Asia (Level 2 in Figure~\ref{fig:batsman_feature_vector}). This classification is significant since the pitch (i.e., the area where the ball is bowled and pitched) conditions vary across the regions, and teams perform differently at different venues~\citep{Vignesh14}. In Section~\ref{Venue_Split_Validationa}, we empirically demonstrate the significance of the difference in scoring patterns at these two classes of venues. For each venue class, the second granularity level contains opposition teams (Level 3 in Figure~\ref{fig:batsman_feature_vector}), divided into $3$ teams' clusters (Section~\ref{teamsclustering}). There are two innings for each match with the opposition, i.e., first and second innings (Level 4 in Figure~\ref{fig:batsman_feature_vector}). From these $3$ granularity levels, we get $12$ different scenarios for $11$ batting performance parameters resulting in a $132$-d feature vector shown in Figure~\ref{fig:batsman_feature_vector}.

The batters clusters are formed with the standard $k$-means clustering on the batters feature vectors (where $k=4$, decided using the standard validation set approach~\citep{validationSetApproach}). The players who never batted are placed into a ``fifth'' cluster. In addition to avoiding the cold-start problem, these batters' clusters are used in match stage feature vector $\Omega(S_i)$ (in Section~\ref{overbyover}) to capture the batters' quality.

\subsubsection{Bowlers Clusters} \label{bowlingclusters}

\begin{figure}[h!]
    \centering
    \includegraphics[scale=0.55]{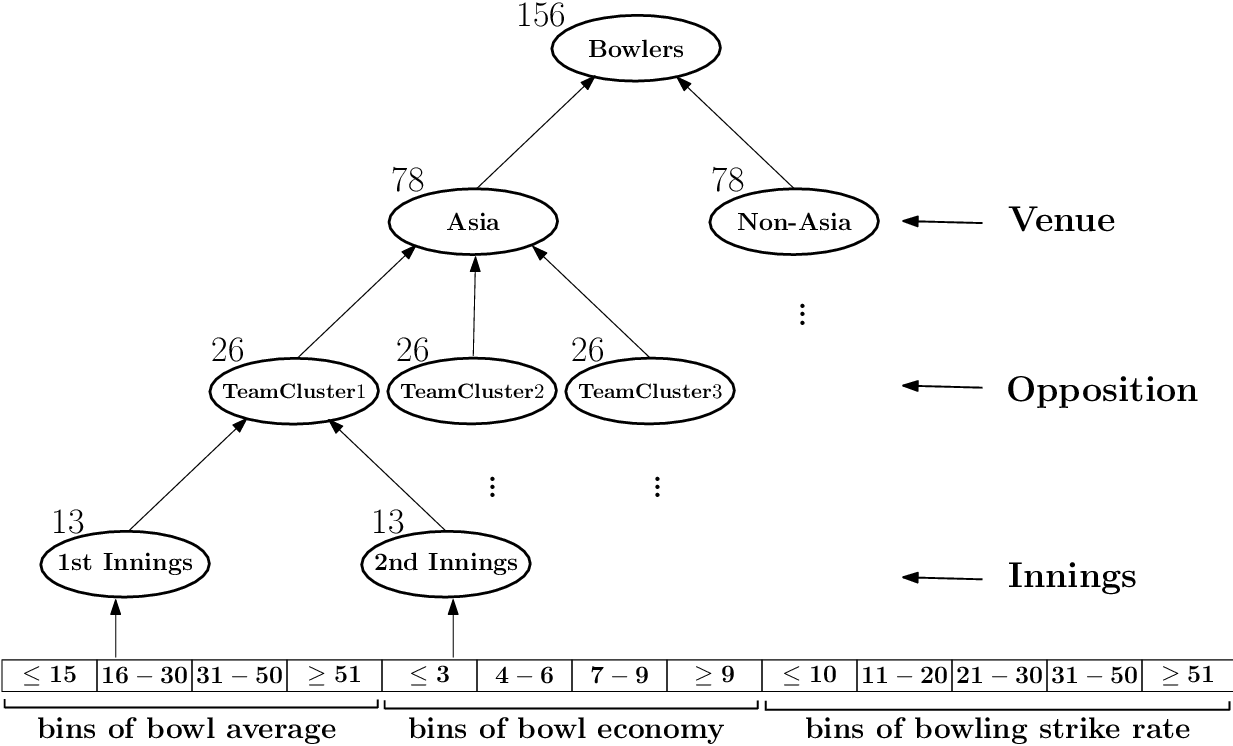}
    \caption{A feature vector for a bowler consists of $4$ bins for bowling average, $4$ bins for economy and $5$ bins for strike rate. The $13$ performance parameters are aggregated across the different venue, opposition, and innings levels to form a $156$-d feature vector.}
    \label{fig:bowler_feature_vector}
\end{figure}

For the representation of bowlers, similar to the batters feature vectors, the bowlers feature vectors contain bowling performance data such as bowling average, strike rate, and bowling economy. The bowling feature vectors, $\psi_q$ for a bowler $q$ contain $13$ bowling performance features, i.e., bowling average, strike rate, and bowling economy discretized into $4$, $5$, and $4$ bins, respectively. 
Similar to batters feature vector, we keep these $13$ performance parameters for the granularity level of innings, opposition's strength, and venue class. 
From these $3$ granularity levels, we get $12$ scenarios for $13$ bowlers performance parameters, resulting in a $156$-d feature vector (Figure~\ref{fig:bowler_feature_vector}). 

The bowlers feature vectors are input to the standard $k$-means clustering algorithm ($k=4$) to obtain the bowlers' clusters. Players having no previous bowling record are placed in a separate ``fifth'' cluster. In addition to avoiding players' cold-start problem, like batters clusters, the bowlers clusters are used in match stage feature vector $\Omega(S_i)$ (in Section~\ref{overbyover}) to capture the bowlers quality.

\subsubsection{Over-by-Over Projections} \label{overbyover}
\begin{figure}[h!]
    \centering
    \includegraphics[scale=0.60]{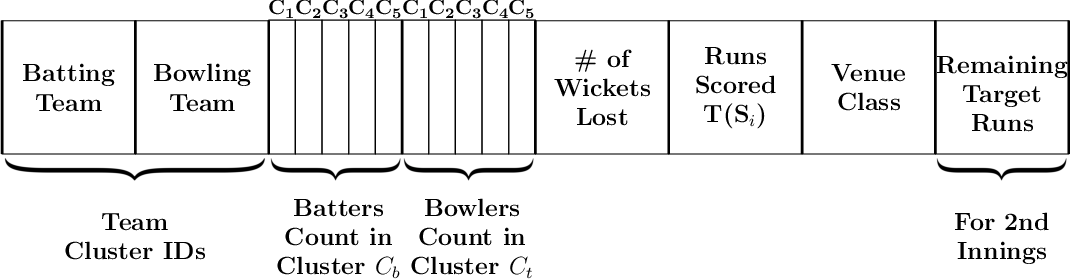}
    \caption{Match stage feature vector $\Omega(S_i)$ for Stage $S_i$ to capture game context, teams, and players' strength. It contains the batting and bowling teams' cluster IDs, cluster-wise counts of remaining batters and bowlers, number of wickets lost, current batting team's score, match venue (Asia and Non-Asia), and remaining target runs.} %
    \label{fig_tbl_state_features} %
\end{figure}

Given the batters and bowlers clusters, we represent the stage of the game by feature vectors, $\Omega(S_i)$ that capture the game context to predict $R(S_i)$ as shown in Figure~\ref{fig_tbl_state_features}.

First, we aggregate ball-by-ball match data to an over-by-over level without loss of necessary information. In $\Omega(S_i)$, cluster ID of batting and bowling teams are used to avoid team sparsity problem. Similarly, to capture the quantitative and qualitative aspect of remaining resources at stage $S_i$, we keep the count of batters $C_b, \;b\in\{1,\cdots,5\}$ belonging to batters cluster $b$ and the count of bowlers $C_t, t\in\{1,\cdots,5\}$ belonging to bowler cluster $t$. The counts of players in respective clusters capture the context in terms of the quality of the players remaining at stage $S_i$, e.g., a stage $S_j$ with $5$ top-order batters of cluster $C_1$ is qualitatively better than the stage $S_k$ with $5$ lower-order batters of cluster $C_4$ and no top-order batters of cluster $C_1$. Further, to quantify bowling resources, we multiply the count of bowlers in $C_t$ with $10$, considering \textsc{odi} rules in which a bowler can bowl a maximum of $10$ overs in an innings. This quantification helps maintain the count of remaining overs and the bowlers' quality. The count of the players who never bowl remains the same across all innings, not affecting the prediction. 

In $\Omega(S_i)$, we also incorporate the match instantaneous features, such as the number of wickets lost, total runs scored, venue class and remaining target runs. This match stage feature vector $\Omega(S_i)$ containing the overall game context is used to predict the expected remaining runs $R(S_i)$ and calculate the players' ratings (\textsc{camp$_{score}$}).

\subsubsection{Projected Score Computation}
For a given $S_i$, we compute the projected total runs in the innings, $P(S_i)$.
The $P(S_i)$ is estimated considering runs scored so far, the number of remaining overs, wickets in hand, the quality of remaining players (batters and bowlers), and the strength of the batting and bowling teams. The teams' strength and players' batting/bowling quality are determined by forming clusters based on their past performance.
The difference between the projected total runs $P(S_i)$ and the total score of a team $T(S_i)$ gives the projected remaining runs $R(S_i)$ for a given $S_i$ in the innings. More formally:

\begin{equation} \label{eq:rsi}
R(S_i) \;=\; P(S_i)-T(S_i)
\end{equation}

We also consider the actual runs scored, $A(S_i)$, by a team after $S_i$. The following section explains the computation of projected remaining runs $R(S_i)$ at any stage of the game. 

\subsubsection{Algorithms for Projected Score Computation}
The main ingredient for $\textsc{camp}_{score}$ is the projected remaining score, $R(S_i)$ at any stage $S_i$ of the game. Algorithm~\ref{algorithmKNN} describes the computation of $R(S_i)$ with the nearest neighbors approach using a test point $\Omega(S_i)'$ feature vector as input. In Line~\ref{algo2ominus}, we use the leave-one-out strategy for the test point $\Omega(S_i)'$ and collect all training examples $\ominus$ corresponding to $S_i$ where wicket lost and overs remaining are equivalent to resources of $\Omega(S_i)'$. In the following line~\ref{algo2aominus}, the actual runs $A{_\ominus}$ for collected training examples $\ominus$ are calculated. We compute the similarity score ($simVec$) using Euclidean distance for the filtered training set (Line~\ref{algo2sim}). In the last line~\ref{algo2weighted}, the target variable $R(S_i)$ is calculated using a weighted average of $simVec$ and $A{_\ominus}$.
\begin{algorithm}[h!]
    \centering
    \caption{$k$NN based projected runs estimation }\label{algorithmKNN}
    \begin{algorithmic}[1]
      \Statex \textbf{Input:}\; $\Omega(S_i)'$    \Comment{Test Point} \label{algo2in}\vskip.03in
      \Statex \textbf{Output:} $R(S_i)$ \label{algo2out}\vskip.03in
       \State $\ominus$ $\gets$ set of $\Omega(S_i)$ with same number of resources as $\Omega(S_i)'$   \Comment{All innings training examples} \label{algo2ominus}\vskip.03in
      \State $A_{\ominus}$ $\gets$ $A$($\Call{Index}{\ominus})$   \Comment{Actual runs vector corresponding to training examples} \label{algo2aominus}\vskip.03in
     \State $simVec \gets \Call{similarity}{\Omega(S_i)',\ominus}$ \label{algo2sim}\vskip.03in
      \State $R(S_i) \gets \Call{WeightedAvg}{simVec \times A_{\ominus}}$ \label{algo2weighted}\vskip.03in
      \end{algorithmic}
    \label{predictionAlgo}
\end{algorithm}

We also compute $R(S_i)$ using regression (Ridge Regression and Random Forest Regressor) with $k$-fold cross-validation, as shown in Algorithm~\ref{algorithmRegression}. We split the input $\Omega(S_i)$ into training and testing sets according to the $k$-fold split (Line~\ref{algo3split}). For each $k$-fold split, we find the indices of the train set (Line~\ref{algo3train}) and test set (Line~\ref{algo3test}). We apply the regression technique to compute our target projected remaining runs vector $R(S_i)$ (Line~\ref{algo3regression}).
\begin{algorithm}[h!]
    \centering
    \caption{Regression-based projected runs estimation} \label{algorithmRegression}
    \begin{algorithmic}[1]
    \Statex \textbf{Input:} $\Omega(S_i)$   \Comment{All matches data for first and second innings}\vskip.03in
    \Statex \textbf{Output:} $R(S_i)$\vskip.03in
    \State $[\Gamma ,\gamma]  \gets \Call{kFoldSplit}{\Omega(S_i)}$ \Comment{$\Gamma$ is train set, $\gamma$ is test set} \label{algo3split}\vskip.03in
     
    \For{j $\gets$ 1:$k$} \vskip.03in
        \State $ind_{tr} \gets \Call{Index}{\Gamma_j}$ \Comment{Indices of train set values} \label{algo3train}\vskip.03in
        \State $ind_{ts} \gets \Call{Index}{\gamma_j}$ \Comment{Indices of test set values} \label{algo3test}\vskip.03in
        \State $R_{ind_{ts}} \gets \Call{Regression}{\Omega(ind_{tr}), A(ind_{tr})  , \Omega(ind_{ts})}$ \label{algo3regression}\vskip.03in
        \Statex \Comment{using Random Forest and Ridge Regression}\vskip.03in
\EndFor
     
    \end{algorithmic}
    \label{regressionAlgo}
\end{algorithm}

\subsection{Computing Players Contributions}
After computation of projected remaining runs $R(S_i)$, our goal is to compute player contributions by \textsc{camp}. $R(S_i)$ and $A(S_i)$ is used to calculate over-by-over contribution scores using expected runs $e_i$ and actual runs $r_i$. These contributions are aggregated for the complete match to obtain all players batting and bowling ratings (\textsc{camp$_{score}$}).

\subsubsection{Estimation of Over-by-over Expected Runs} \label{expectedrunsestimation}
After computation of projected remaining runs $R(S_i)$ for a given $S_i$, we compute the expected runs for $i^{th}$ over, $e_i$. The change between $R(S_i)$ and $R(S_{i+1})$ is equivalent to expected runs, $e_i$, in the $i^{th}$ over. More formally: 

\begin{equation}
e_i \;=\; R(S_i) -R(S_{i+1})   \label{eq:expectedrunsOver}
\end{equation}

There are two possible scenarios in an over $i$: either the batting team loses wicket(s) or not. The expected runs for that over, $e_i$, change accordingly. In case of the wicket(s) lost in an over, the team's capability to score runs in the remaining part of the innings is affected, and $P(S_{i+1})$ decreases depending on the importance of the wicket lost. As a result, the change in two projections $P(S_i)$ and $P(S_{i+1})$ increases compared to the case when no wicket is lost. This increased difference in $P(S_i)$ and $P(S_{i+1})$ is due to the higher worth of the wicket lost, i.e., if wicket(s) is/are lost in initial overs, the change in two projections will be higher than that if the wicket is lost in final overs.

For each wicket lost, $e_i$ is modified according to wicket weight ($w$) to penalize the outgoing batter and reduce the expectation from the incoming batter. $e_i$ remains the same for no loss of wicket. More formally:
\begin{equation}
  e_i' \;=\; \begin{cases}
      (1-w)e_i & \quad \text{wicket lost, $w\in[0.1,1]$} \\
   e_i   & \quad  \text{otherwise } %
   \label{eq:expectedRunsOverCasesWicketLost}
  \end{cases}
\end{equation}

These expected runs, $e_i'$ are used to calculate players' contributions in equation~\eqref{eq:batsmanContributionInOver} and equation~\eqref{eq:bowlerContributionInOver} of Section~\ref{playersratingformulation}.

\begin{remark}
Note that \textsc{MoM} is an expert opinion-based metric and identifies the {\em ``top performing''} player. We use it to validate the players' rating computed by \textsc{camp}. 
\end{remark}
\textsc{MoM} is the only metric that provides a baseline measure to compare the top contributor of \textsc{camp}. Therefore, wicket weights ($w$) are adjusted empirically by maximizing the agreement of the top contributor by \textsc{camp} with the experts' opinion-based top contributor (\textsc{MoM}). We use a varying value of $w$ to get a maximal agreement of our top contributor with the \textsc{MoM}. We use $w \in [0.1,1]$ with the increase of $0.05$ and for $w =1$, we get maximum matching with \textsc{MoM}.
The selection of $w$ is not a subjective decision. $w$ serves as a hyperparameter of our technique, which is not required to be adjusted for each iteration. To bring the expectation level to ball-by-ball, $e_i'$ is uniformly divided among each ball of the over as $\nicefrac{e_i'}{6}$.
\subsubsection{Computing Over-by-over Contribution Scores} \label{playersratingformulation}
As the innings proceeds, we compute $R{(S_i)}$, the projected remaining runs in the innings. We also consider the actual runs scored, $A(S_i)$, by a team after $S_i$. Thus, the actual runs scored in over $i$ are as follows: %
\begin{equation}
r_i \;=\; A(S_{i}) - A(S_{i+1}) \label{eq:actualRunsOver}
\end{equation}
Similarly, $r_i^{p}$ represents the actual runs scored by batter $p$ in over $i$, where $p \in [1,22]$ is the unique identifier for each player. For a batter facing the bowler, his contribution is quantified by how well he performs with respect to $e_i'$. The expected score for a batter $p$ is computed as $\nicefrac{e_i'\times b_p}{6}$, where $b_p$ is the number of balls faced by the batter in the respective over (recall that an over consist of $6$ balls). The contribution $c_i^p$ of the batter $p$ in $i^{th}$ over is computed as follow:
\begin{equation}
c_i^{p} \;=\; r_i^{p} - \dfrac{e_i'}{6}
\times b_p\; \quad p \in[1,22] %
\label{eq:batsmanContributionInOver}
\end{equation}
The net contribution in $i^{th}$ over ($c_i^p$) can be positive or negative depending on whether the batter scored above or below expectation. A positive batter contribution implies a negative contribution of the bowler and vice versa. Similarly, minimizing the runs conceded in an over or taking wickets contribute positively towards the bowler's contribution.

\begin{remark}
Note that batters are only credited for the runs they score but for a bowler's extras (e.g., wide ball, no ball) are also counted as runs conceded by the bowler.
\end{remark}

The contribution of a bowler is computed as:
\begin{equation}
\hat{c}_i^{p} \;=\;  e_i' - r_i \; \quad p \in[1,22]
\label{eq:bowlerContributionInOver}
\end{equation}
\subsubsection{Computing Players Rating using Over-by-over Contribution Vector}
After computing over-by-over contribution scores of players for both innings of a match, we aggregate contributions $c_i^{p}$ and $\hat{c}_i^{p}$ over a complete match for each player. Since both teams have $11$ players, we associate batting and bowling contributions with each player to get a $44$-d resultant vector.

If a batter remains on the crease for overs in a set $Q$ and loses his wicket in $j^{th}$ over, his aggregated batting contribution is computed as:
\begin{equation}
  C_{bat}(p) \;=\;\begin{cases}
     \sum_{i \in Q }   {  c_i^{p} - (w \times e_j)}& \text{wicket lost} \\
   \sum_{i \in Q}     c_i^{p} &  \text{otherwise} %
   \label{eq:battingContribution}
  \end{cases}
\end{equation}
For a bowler, who bowled overs in a set $Q$, his contribution is defined analogously as:
\begin{equation}
  C_{bowl}(p) \;=\;
     \sum_{i \in Q } \hat{c}_i^{p} + \sum_{k\, \in \text{ overs with wickets}} (w \times e_k)
    \label{eq:bowlingContribution}
\end{equation}
 
\begin{remark}
A wicket loss by run-out is debited against the batter but is not credited to the bowler.
\end{remark}
We compute the net contribution, \textsc{camp$_{score}$} (players' rating) as follows:
\begin{equation}
{\textsc{camp$_{score}$}} \;=\;  w_{bat} \times C_{bat}(p)+ w_{bowl} \times C_{bowl}(p)
\label{netContribution}
\end{equation}

where $w_{bat}$ and $w_{bowl}$ are user-set parameters and weight batting and bowling contributions, respectively. We use varying weights for batting and bowling contributions in Equation~\eqref{netContribution} to calculate all players' ratings as \textsc{camp$_{score}$} vector. To make a comparison with \textsc{MoM}, we adjust weights ($w_{bat}$ and $w_{bowl}$) such that the top contributor from \textsc{camp} agrees with \textsc{MoM}. For $w_{bat}=1$ and $w_{bowl}=0.2$, we get maximum matching with the expert opinion-based top contributor \textsc{MoM}. The players' contribution scores can be aggregated to match, series, or tournament level along multiple dimensions (e.g., batting, bowling, or both). Using aggregate contribution scores of each player, the competitive balance of a cricket competition at any level can be assessed by analyzing the distribution of aggregate contribution scores across teams. A high level of parity between the teams suggests a relatively even distribution of scores across all teams which helps to improve the balance and overall quality of the competition. This paper shows our work at the match and series level; however, the approach can be extended to any level.

\subsubsection{The \textsc{camp} Algorithm} 
Algorithm~\ref{CAMPalgorithm} contains the pseudo-code to compute \textsc{camp$_{score}$} vector for all $22$ players. It uses Algorithm~\ref{predictionAlgo} or Algorithm~\ref{regressionAlgo} as a subroutine to project the remaining score at a stage. In Line~\ref{algo1cbatt} and Line~\ref{algo1cbowl}, we respectively form the batters and bowlers clusters $\lambda_{batt}$ and $\lambda_{bowl}$, using batters and bowlers feature vectors $\phi(\cdot)$ and $\psi(\cdot)$. In Line~\ref{algo1omegasi}, we use the batters and bowlers clusters along with instantaneous match features at match stage $S_i$ to obtain the match stage feature vector, $\Omega(S_i)$. Line~\ref{algo1rsi} computes projected remaining score at stage $S_i$, $R(S_i)$ using $\Omega(S_i)$ (Algorithm~\ref{predictionAlgo}). In Line~\ref{algo1campscore}, \textsc{camp$_{score}$} is calculated from $R(S_i)$ and the actual runs data $A(S_i)$ by Equation~\eqref{netContribution}.
\begin{algorithm}[h!]
    \centering
    \caption{\textsc{camp} algorithm for players ratings}
    \begin{algorithmic}[1]
      \Statex \textbf{Input:}\; Batters Data$~\phi$, Bowlers Data$~\psi$, Ball-by-Ball Data $A$ \label{algo1in}\vskip.03in
      \Statex \textbf{Output:}  Players Ratings (\textsc{camp$_{score}$}) \label{algo1out}\vskip.03in
      \State $\lambda_{batt} \gets$ \Call{PerformClustering}{$\phi$} \Comment{$k$-means with $k=4$, Section~\ref{battingclusters}}  \label{algo1cbatt}\vskip.03in
      \State $\lambda_{bowl} \gets $   \Call{PerformClustering}{$\psi$} \Comment{$k$-means with $k=4$, Section~\ref{bowlingclusters}} \label{algo1cbowl}\vskip.03in
      \For{$i= 1 \to 50$}\vskip.03in
      \State $\Omega(S_i) \gets$ \Call{GenerateFeatureVector}{$S_i$, $\lambda_{batt}$, $\lambda_{bowl}$} \Comment{Section~\ref{overbyover}}  \label{algo1omegasi}\vskip.03in
      \State $R(S_i) \gets $ \Call{EstimateProjection}{$\Omega(S_i)$} \Comment{ Section~\ref{expectedrunsestimation}}  \label{algo1rsi} \vskip.03in
      \State \textsc{camp$_{score}$} $\gets$ \Call{ComputeRatings}{$R(S_i)$, $A(S_i)$}  \label{algo1campscore} \Comment{Section~\ref{playersratingformulation}}\vskip.03in
      \EndFor
    \end{algorithmic}
    \label{CAMPalgorithm}
\end{algorithm}

\section{Experimental Setup}\label{experimentalSetup}
This section describes our dataset consisting of one-day international cricket matches and players, along with preprocessing of the dataset. Moreover, we discuss the performance metrics used to evaluate the proposed model against baseline methods.

\subsection{Dataset Statistics}
ESPNcricinfo\footnote{\url{https://www.espncricinfo.com/}}, a leading sports website, records cricket data for every match played under the ICC rules. We extracted ball-by-ball data, match summaries, and player performance statistics at the innings level from ESPNcricinfo. We used the data of $1625$ complete \textsc{odi} matches played between January $2001$ to October $2019$ among $10$ full-time ICC member teams (Table~\ref{tab_teamsCluster}) in our analysis.

\subsubsection{Players' Data}
The individual players' data comprises performance statistics aggregated to the innings level for all matches. The players' performance data is divided into batting and bowling data. Batters data consists of $1002$ unique players from the top $10$ teams who faced at least one ball while bowling data contains $802$ unique bowlers who have bowled at least one over in their \textsc{odi} career.
We have made this comprehensive preprocessed dataset and our code publicly available online\footnote{\url{https://github.com/sohaibayub/CAMP}} for academic research.

\subsubsection{Match Summary Data}
The match summary data contains the general and specific information of participating teams, venue, date, toss-winner, total runs scored in both innings, wickets lost, run rates, match winner, and victory margin, respectively. The total runs scored in any innings show the team's batting capability and the bowling strength of the opposition. The most important piece of information in match summary data is the player declared as Man of the Match (\textsc{MoM}), which we use to validate $\textsc{camp}_{score}$.

\subsection{Data Preprocessing}
\begin{figure}[h!]
\centering
\begin{subfigure}{.45\textwidth}
    \centering
    \includegraphics[width=.9\linewidth]{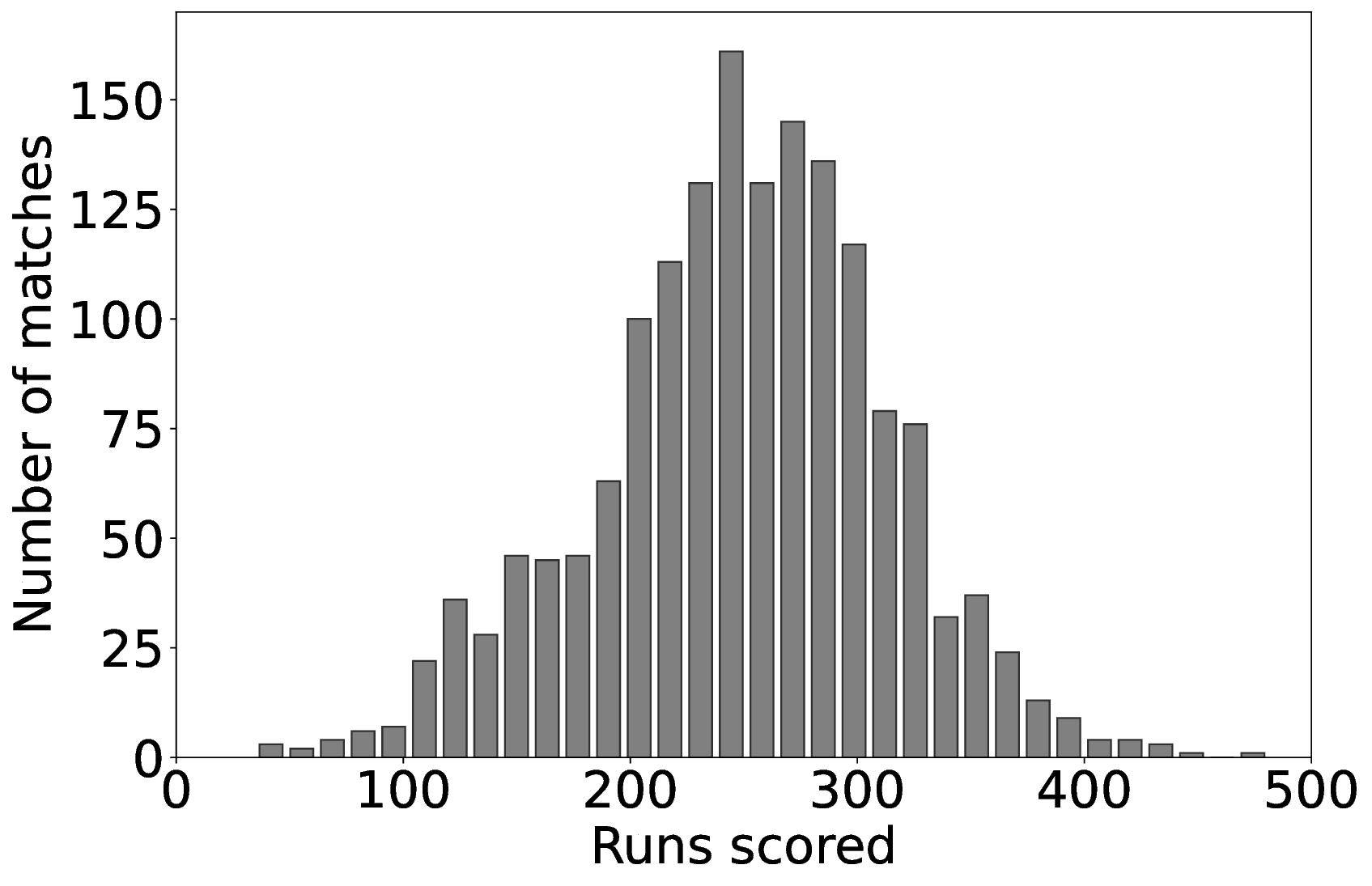}
    \caption{First innings before preprocessing}
    \label{figure_innings_1_before}
\end{subfigure}%
\begin{subfigure}{.45\textwidth}
  \centering
    \includegraphics[width=.9\linewidth]{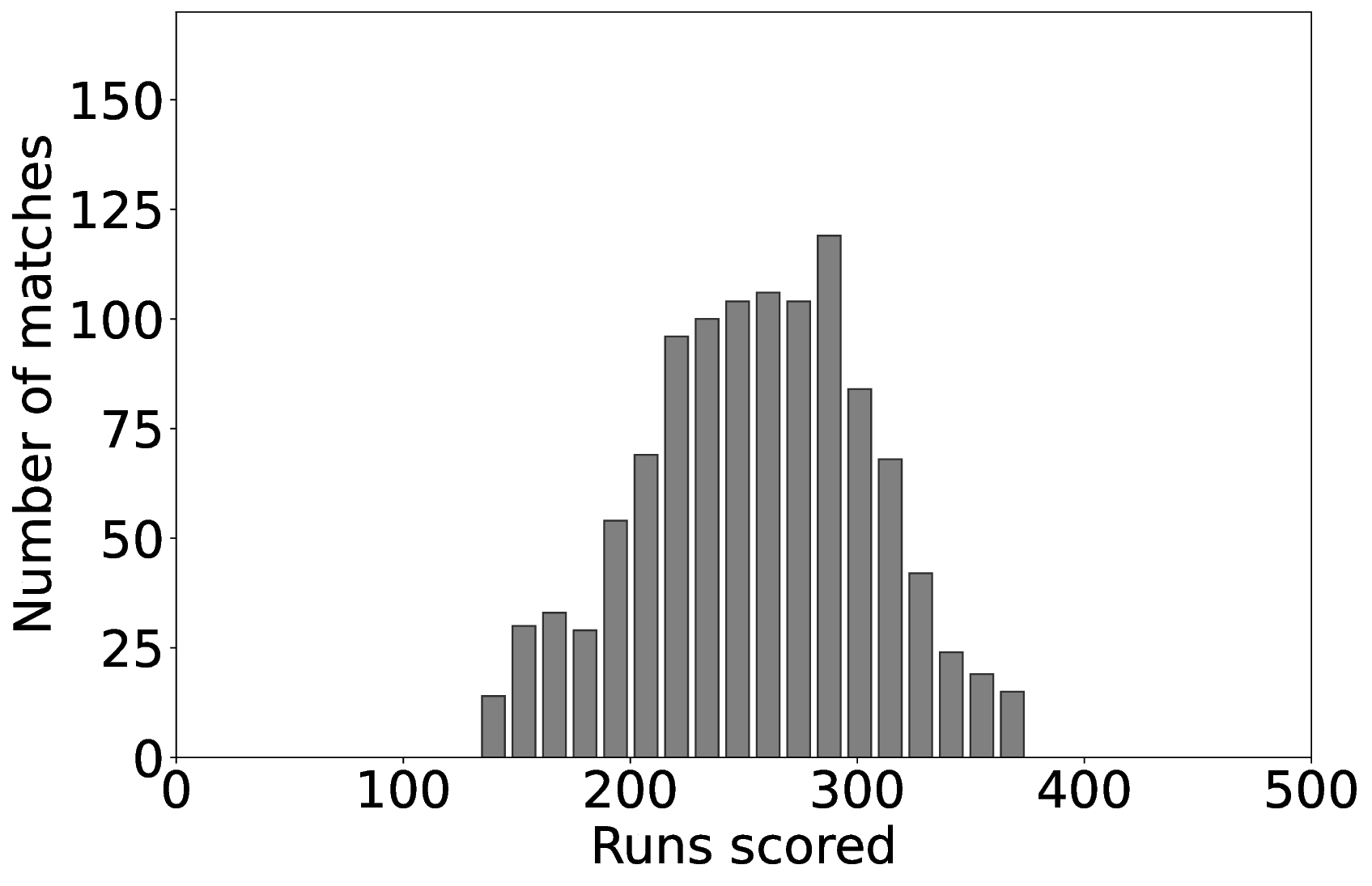}
    \caption{First innings after preprocessing}
    \label{figure_innings_1_after}
\end{subfigure} \\
\begin{subfigure}{.45\textwidth}
    \centering
    \includegraphics[width=.9\linewidth]{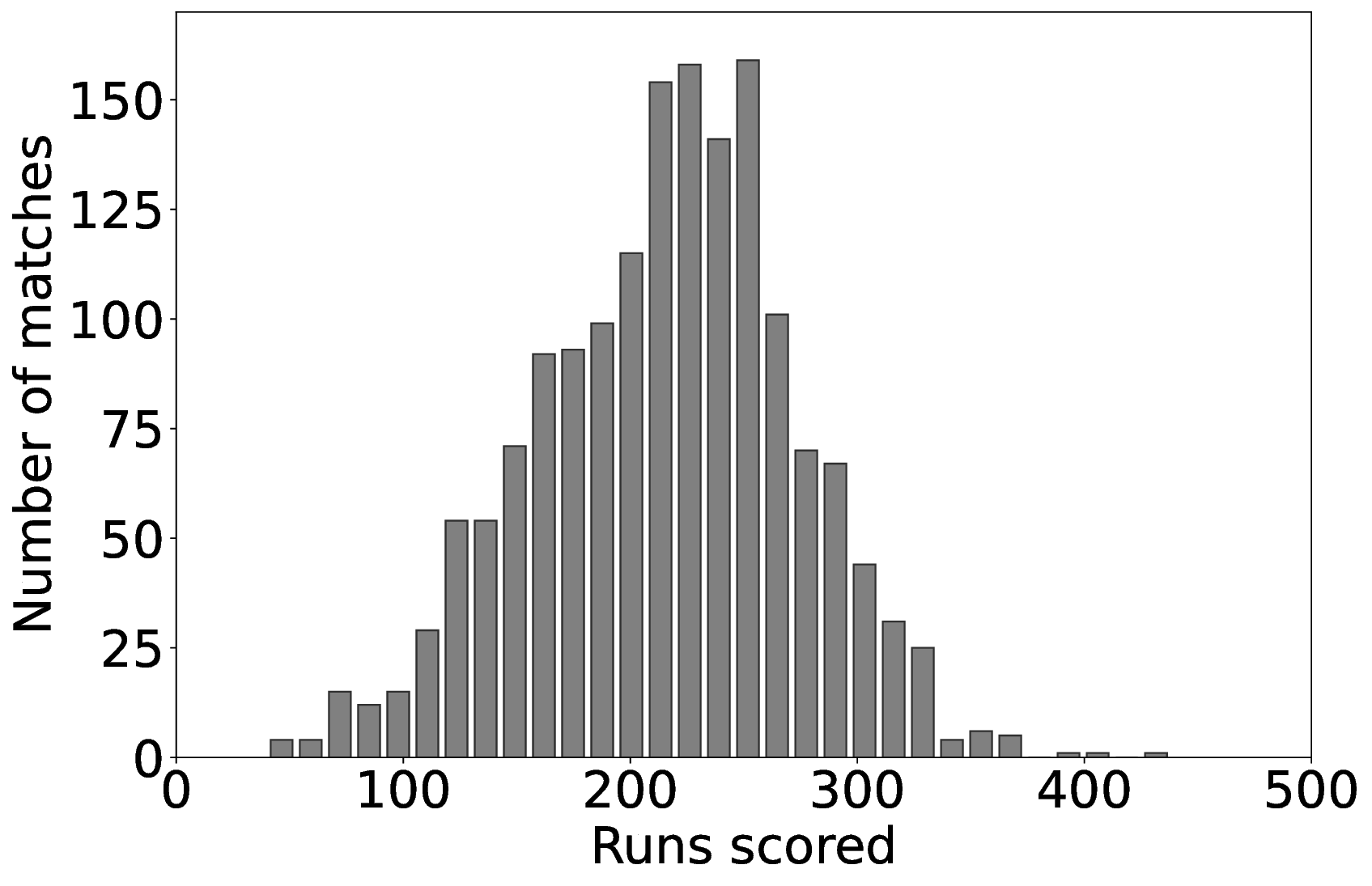}
    \caption{Second innings before preprocessing}
    \label{figure_innings_2_before}
\end{subfigure}%
\begin{subfigure}{.45\textwidth}
  \centering
    \includegraphics[width=.9\linewidth]{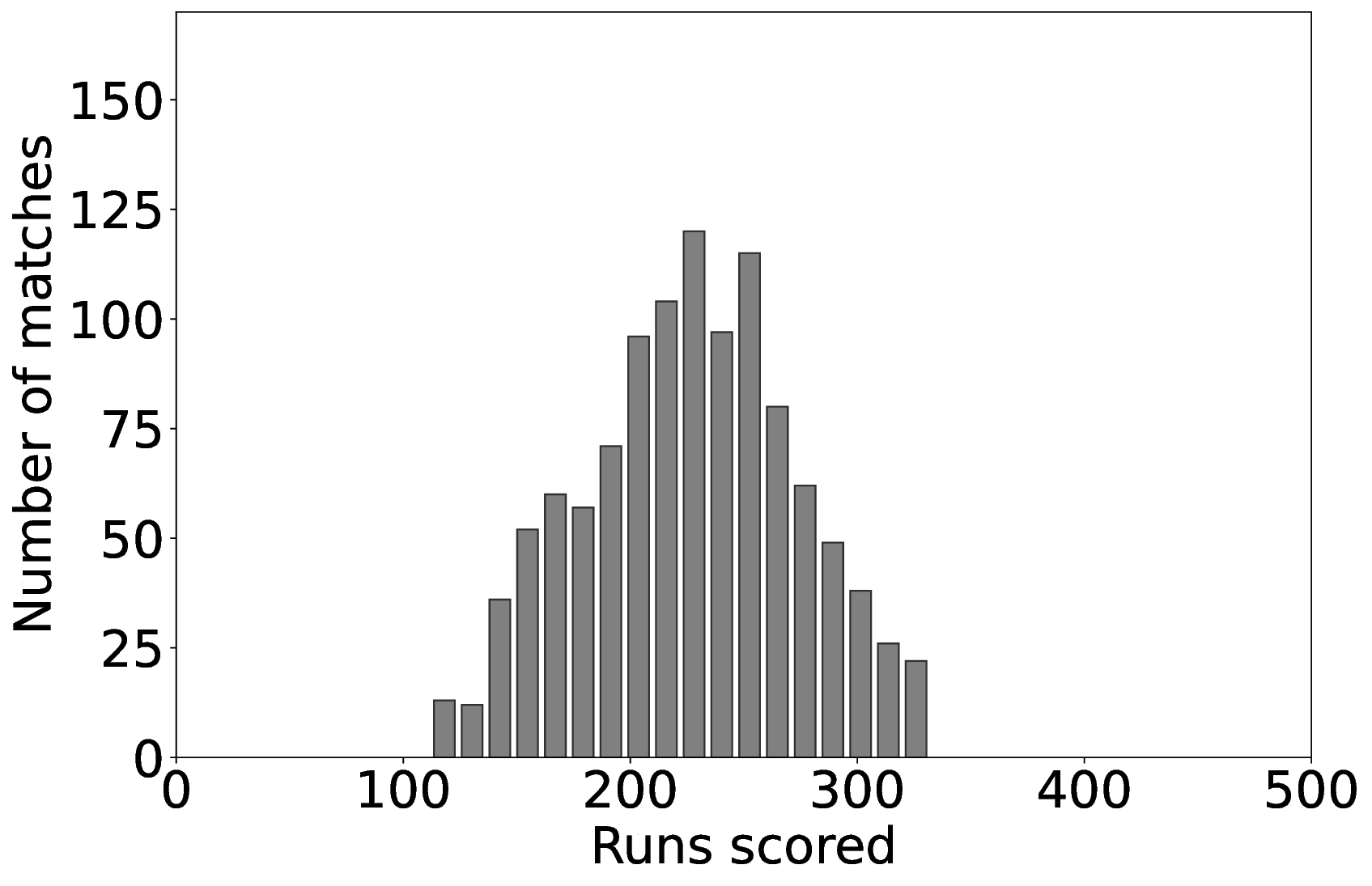}
    \caption{Second innings after preprocessing}
    \label{figure_innings_2_after}
\end{subfigure}
\caption{Total runs distribution of all matches in both innings before and after removing the low runs scorer teams (BAN and ZIM) and matches with runs scored less than two standard deviations from mean runs.}
\label{fig_innings_preprocessing}
\end{figure}

We preprocess the data to remove inconsistencies and find the most informative set of matches. We only keep those matches in which the runs scored in both innings are within $2$ standard deviations of the mean innings scores. We observe that the two teams, BAN and ZIM (with lower ICC rankings during the sampled years), generally scored significantly less than other teams. We removed all matches involving these two teams. Figure~\ref{fig_innings_preprocessing} shows the distributions of innings scores before and after removing outliers. A summary of match scores before and after preprocessing is given in Table~\ref{tab:Innings_total_statistics}. 

\begin{table}[h!]
    \centering
    \small
    \begin{tabular}{@{\extracolsep{4pt}}lccccc@{}}
      \toprule
     & \multicolumn{2}{c}{{All 1625 matches}} & &
    \multicolumn{2}{c}{{After preprocessing $1110$ matches}} \\
    \cmidrule{2-3} \cmidrule{5-6}
     & { First Innings} & { Second Innings }  & & { First Innings} & {Second Innings } \\
     \midrule
       {Min} & $35$ & $40$&& $133$ &  $112$\\
       {Max} & $481$ & $438$&& $375$  & $332$ \\
       {Mean} &  $249$ & $216$&& $256$ & $226$ \\
       {Std.} & $64$  & $58$ &\qquad & $50$ & $47$ \\
       \bottomrule
    \end{tabular}
    \caption{Statistics of runs for both innings before and after removing outlier matches, i.e., the matches with average runs scored beyond two standard deviations from mean runs and matches played by the low-scoring teams (BAN and ZIM).}
    \label{tab:Innings_total_statistics}
\end{table}

\subsection{Evaluation Measures}
We evaluate the effectiveness of \textsc{camp} in terms of accuracy of the projected scores, quality of players' ratings, and by validating the teams and players clustering. We compare the projected scores $R(S_i)$ by $k$NN, Random Forest, and Ridge Regression with the actual runs scored $A(S_i)$ and report the mean absolute error (\textsc{mae}). We also report the \textsc{mae} of $R(S_i)$ computed by \textsc{lnc} based on the resource table in~\citep{Lewis05}. For \textsc{lnc}, we use the publicly available Duckworth-Lewis (\textsc{dl}) resource table (Table~\ref{tab:DLS_Table} in Appendix). \textsc{lnc} proposes $Z(50,0)=235$ for the first innings and target runs for the second innings as expected runs with all wickets in hand and $50$ overs remaining. The table entries show the percentage of $Z$ runs that can be scored after a specific stage. 

We can only evaluate players' performance based on the agreement of the top contributor (top-rated player) of \textsc{camp} with the \textsc{MoM} declared by the ICC since there is no ground truth for players' true contributions in a given match. We report the fraction of matches in which \textsc{MoM} is the top and one of the top contributors by \textsc{camp}. We also compare the \textsc{camp} ratings with \textsc{lnc} both at the match and series levels.

We also validate the intermediate steps of teams' and players' clustering to demonstrate that our feature vectors  are meaningful and that the clusterings are well-formed.

\section{Results and Discussion}\label{resultsAndDiscussion}
In this section, we start with validating players' clusters using ICC's top 100 players' ratings for bowling and batters clusters. We show that these are well-formed quality clusters using clusters of top ICC-rated players in Section~\ref{playerClusterValidation} and visually using t-SNE diagrams in Section~\ref{playerstsne}. In the next Section~\ref{shapanalysis}, we investigate the important features from the players' feature vector. Section~\ref{Venue_Split_Validationa} explains the validation for venue-wise distribution of teams. We perform the evaluation of \textsc{camp} using projected remaining runs and players' ratings in Section~\ref{projectedrunsevaluation} and Section~\ref{ratingevalution}, respectively.

\subsection{Players' Clustering Validation using ICC Ratings}\label{playerClusterValidation}
We compare the players' clustering with the ICC top players' rankings to evaluate the goodness of batters and bowlers clusters. The historical data for players' clustering from January 1, 2000 to October 20, 2019 along with the ICC top players rankings on October 20, 2019\footnote{\href{http://www.relianceiccrankings.com/datespecific/odi/batting/2019/10/20/}{ICC Men's \textsc{odi} Players Rankings on October 20, 2019} - \url{https://www.icc-cricket.com/rankings/mens/player-rankings/odi?at=2019-10-20}} is used for clustering validation. 
Table~\ref{tab:batsmenCluster} shows the batters and bowlers clusters for ICC's top $10$ players. All ICC top-ranked batters are in the same batters cluster, validating the quality of our batters clusters. Whereas the bowlers clusters of these batters vary as opposed to the batters cluster showing that the top-ranked batters do not necessarily have the same bowling quality. For example, a few batters (e.g., B. Azam, Q. Kock and J. Roy) are in the fifth dummy bowlers cluster as they have never bowled. Similarly, ICC's top $10$ bowlers belong to the two nearby clusters of bowlers. Moreover, clusters containing top batters are generally mutually exclusive with clusters containing top bowlers except for the case of all-rounders. For example, ``C. Woakes", a good all-rounder, is in the same cluster $2$ as the top $10$ ICC batters in Table~\ref{tab:batsmenCluster}.

\begin{table}[h!]
    \centering
	\begin{minipage}[t]{0.48\textwidth}
    \vspace{0pt}
    \resizebox {1\textwidth} {!} {%
		\begin{tabular}{clccc}
			\toprule
			\TwoRows{ICC Batter} {Rank} & {Name} & \TwoRows{ICC}{Rating} & \TwoRows{Batters}{Cluster} & \TwoRows{Bowlers}{Cluster} \\
			\midrule
			1 & V. Kohli & $895$& $2$ & $1$\\
			2 & R. Sharma & $863$ & $2$ & $1$\\
			3 & B. Azam & $834$ & $2$ & $5$\\
			4 & F. Plessis & $820$& $2$ & $1$\\
			5 & L. Taylor & $817$ & $2$ & $2$\\
			6 & K. Williamson & $796$ & $2$ & $2$\\
			7 & D. Warner & $794$& $2$ & $1$\\
			8 & J. Root & $787$ & $2$ & $1$\\
			9 & Q. Kock & $781$ & $2$ & $5$\\
			10 & J. Roy & $774$& $2$ & $5$\\
			\bottomrule
		\end{tabular}%
    }
    \end{minipage}
    \quad
    \begin{minipage}[t]{0.47\textwidth}
    \vspace{0pt}
    \resizebox {1\textwidth} {!} {%
		\begin{tabular}{clccc}
			\toprule
			\TwoRows{ICC Bowler}{Rank} & {Name} & \TwoRows{ICC}{Rating} & \TwoRows{Bowlers} {Cluster} & \TwoRows{Batters} {Cluster}\\
			\midrule
			1 & J. Bumrah & $797$ & $3$& $4$\\
			2 & T. Boult & $740$  & $3$& $1$\\
			3 & K. Rabada & $694$  & $3$& $1$\\
			4 & P. Cummins & $693$ & $4$& $1$\\
			6 & C. Woakes & $676$  & $3$& $2$\\
			7 & M. Starc & $663$  & $4$& $4$\\
			7 & M. Amir & $663$ & $3$& $1$\\
			8 & M. Henry & $656$  & $4$& $4$\\
			9 & L. Ferguson & $649$  & $4$& $4$\\
			10 & K. Yadav & $642$ & $3$& $1$\\
			\bottomrule
		\end{tabular}%
    }
    \end{minipage}	
	\caption{ICC top-ranked batters and bowlers with their cluster IDs. All top-ranked players are grouped into the same or nearby clusters showing that clustering captures the players' quality. Top all-rounders (e.g., C. Woakes) belong to the top-quality batters and top-quality bowlers cluster.}%
    \label{tab:batsmenCluster}
\end{table}

\subsection{Players' Clustering Validation using Feature Vectors Visualization} \label{playerstsne}

\begin{figure}[h!]
\centering
   \begin{subfigure}[b]{0.45\textwidth}
   \includegraphics[width=0.95\textwidth]{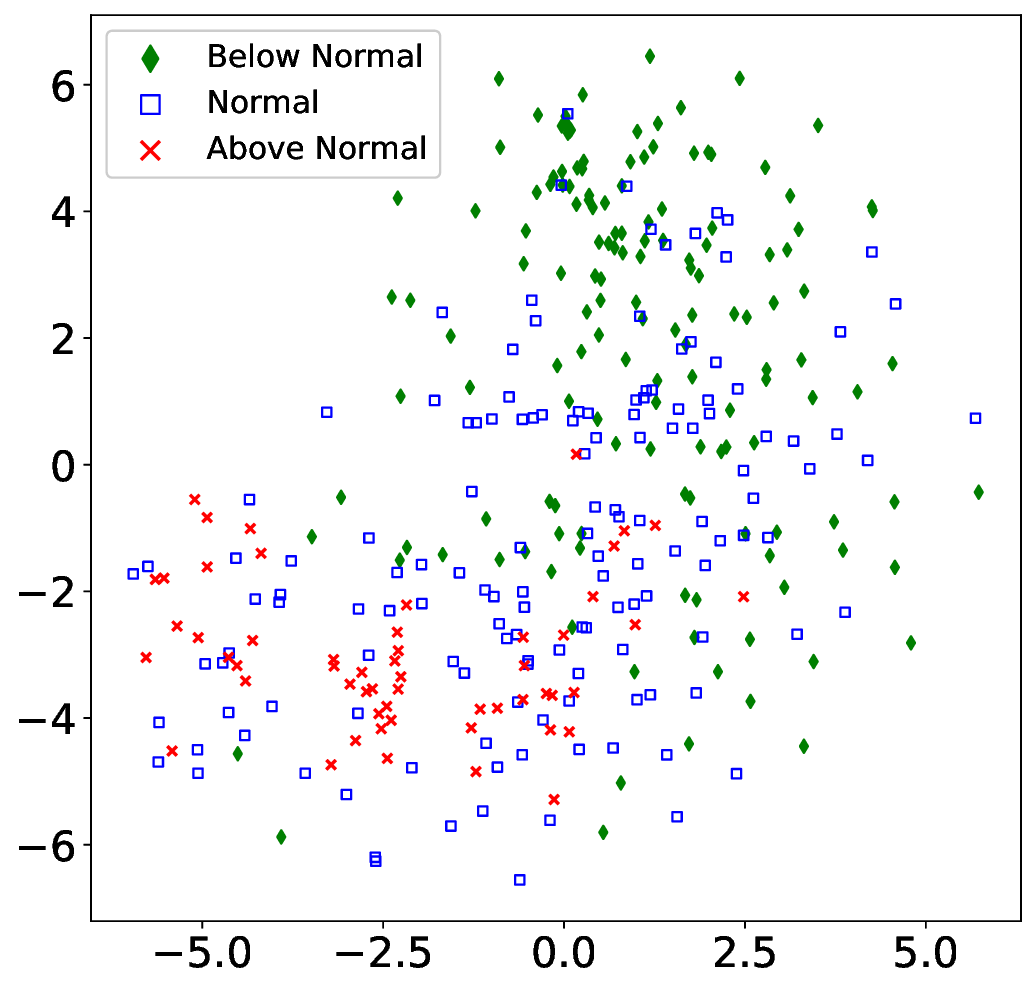}
   \caption{Batters}
   \label{fig_tsne_batsmen} 
   \end{subfigure}
   \begin{subfigure}[b]{0.45\textwidth}
   \includegraphics[width=0.95\textwidth]{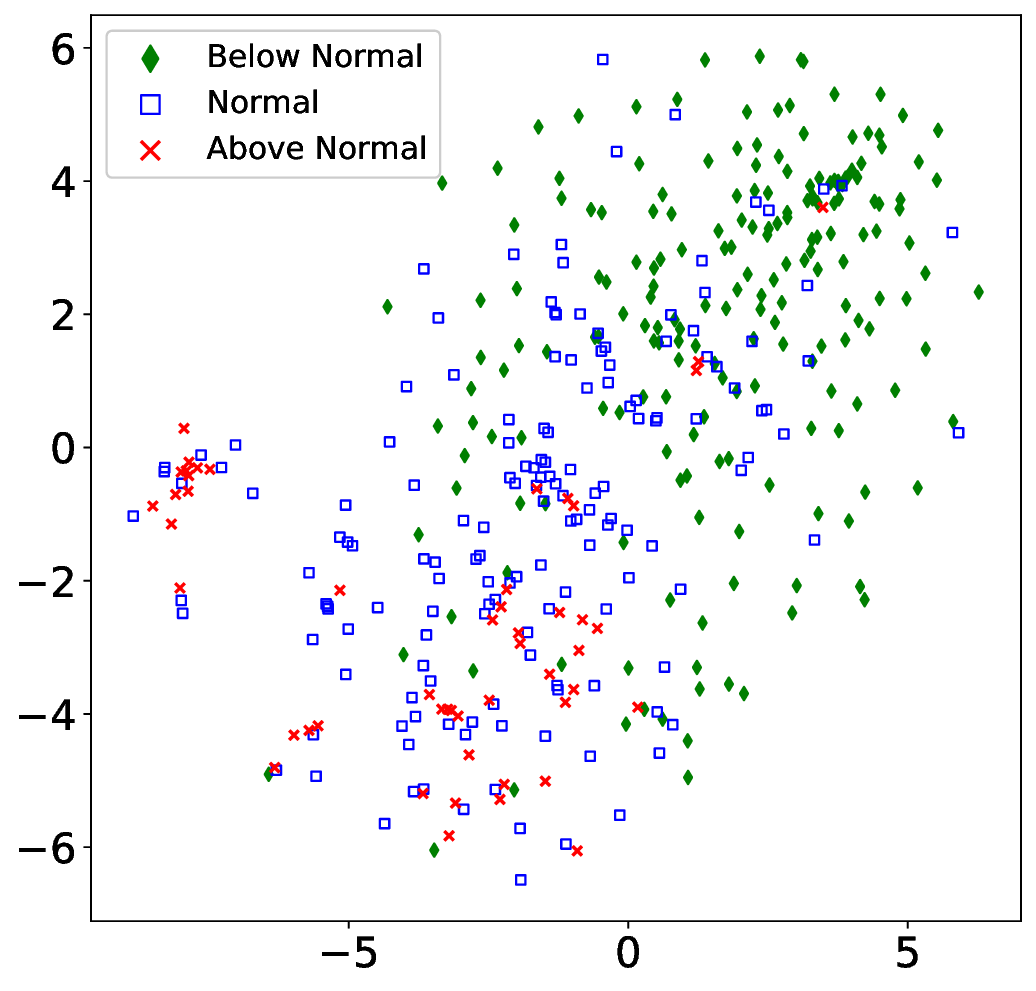}
   \caption{Bowlers}
   \label{fig_tsne_bowlers}
\end{subfigure}
    \caption{$156$-d and $132$-d batters and bowlers feature vectors mapped to $\mathbb{R}^2$ using $t$-SNE in \textbf{(a)} and \textbf{(b)}, resp.. The aggregated ICC quarterly players ratings from $2001$ to $2019$ are used as labels to group similarly rated players. Figures are best seen in color.} %
    \label{tsne2DAllFeats}
\end{figure}

To visualize the batters and bowlers feature vectors, we use $t$-distributed stochastic neighbor embedding ($t$-SNE)~\citep{van2008visualizing} to map the data into $\mathbb{R}^2$ (Figure~\ref{tsne2DAllFeats}). We collected the quarterly ICC player ratings of the top $100$ batters and bowlers from 2001 to 2019 (total $76$ measurements). These ratings are aggregated for each player giving a total of $410$ ICC-rated batters and $376$ bowlers, i.e., the players rated at least once from 2001 to 2019. These aggregate ratings, grouped into three clusters (using $k$-means with $k=3$), are used as labels for players' feature embeddings in the $t$-SNE diagram.
We observe that the players with similar ICC ratings lie in the same proximity in the t-SNE diagram (Figure~\ref{tsne2DAllFeats}). This demonstrates that the players' feature vectors capture the players' quality (determined by the ICC's top players' ratings). 

\subsection{SHAP Analysis for Players' Feature Importance} \label{shapanalysis}
\begin{figure}[h!]
\centering
   \begin{subfigure}[b]{0.49\textwidth}
   \includegraphics[width=\textwidth]{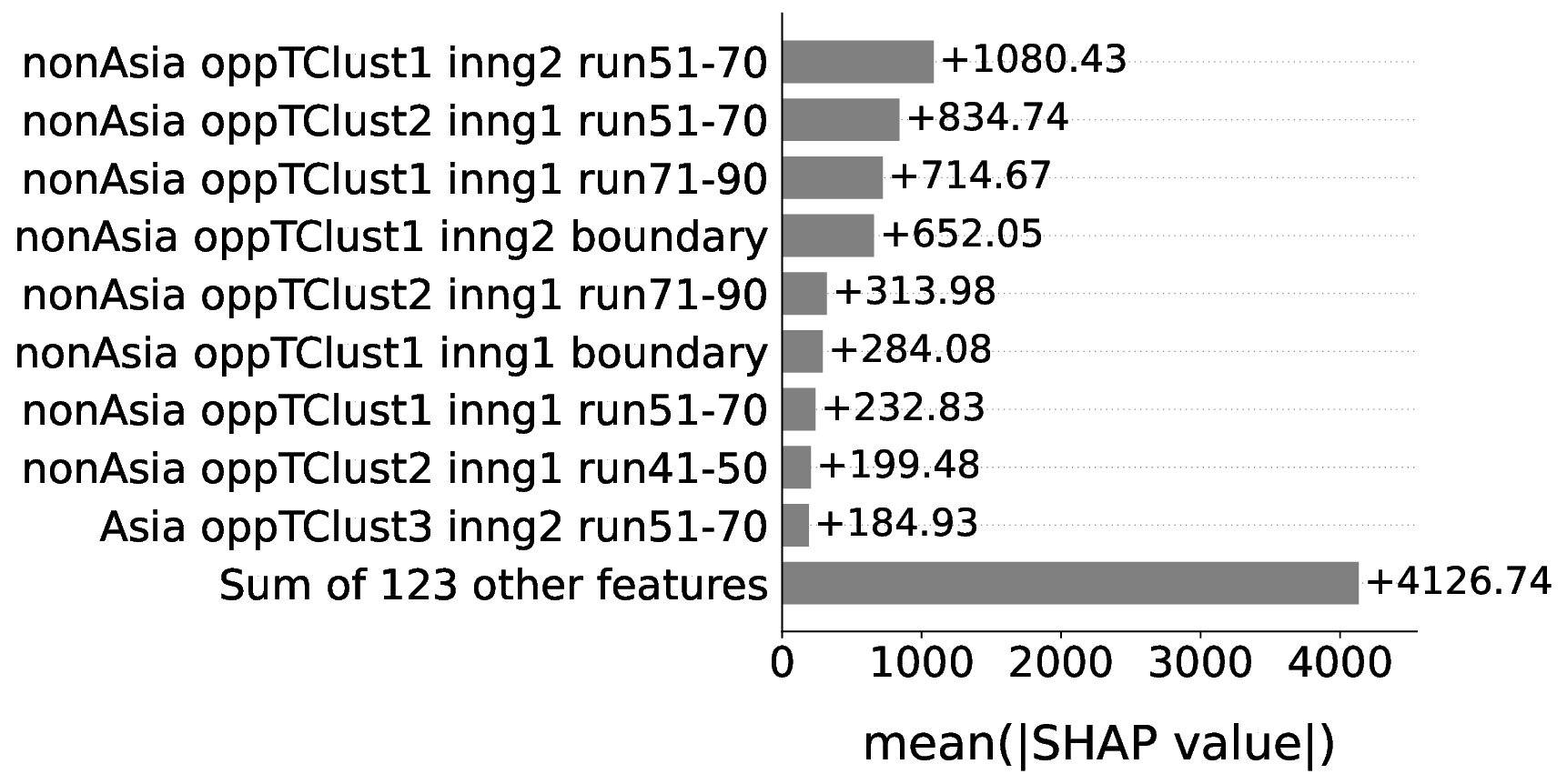}
   \caption{Batters}
   \label{fig_shap_batsmen} 
   \end{subfigure}
   \begin{subfigure}[b]{0.49\textwidth}
   \includegraphics[width=\textwidth]{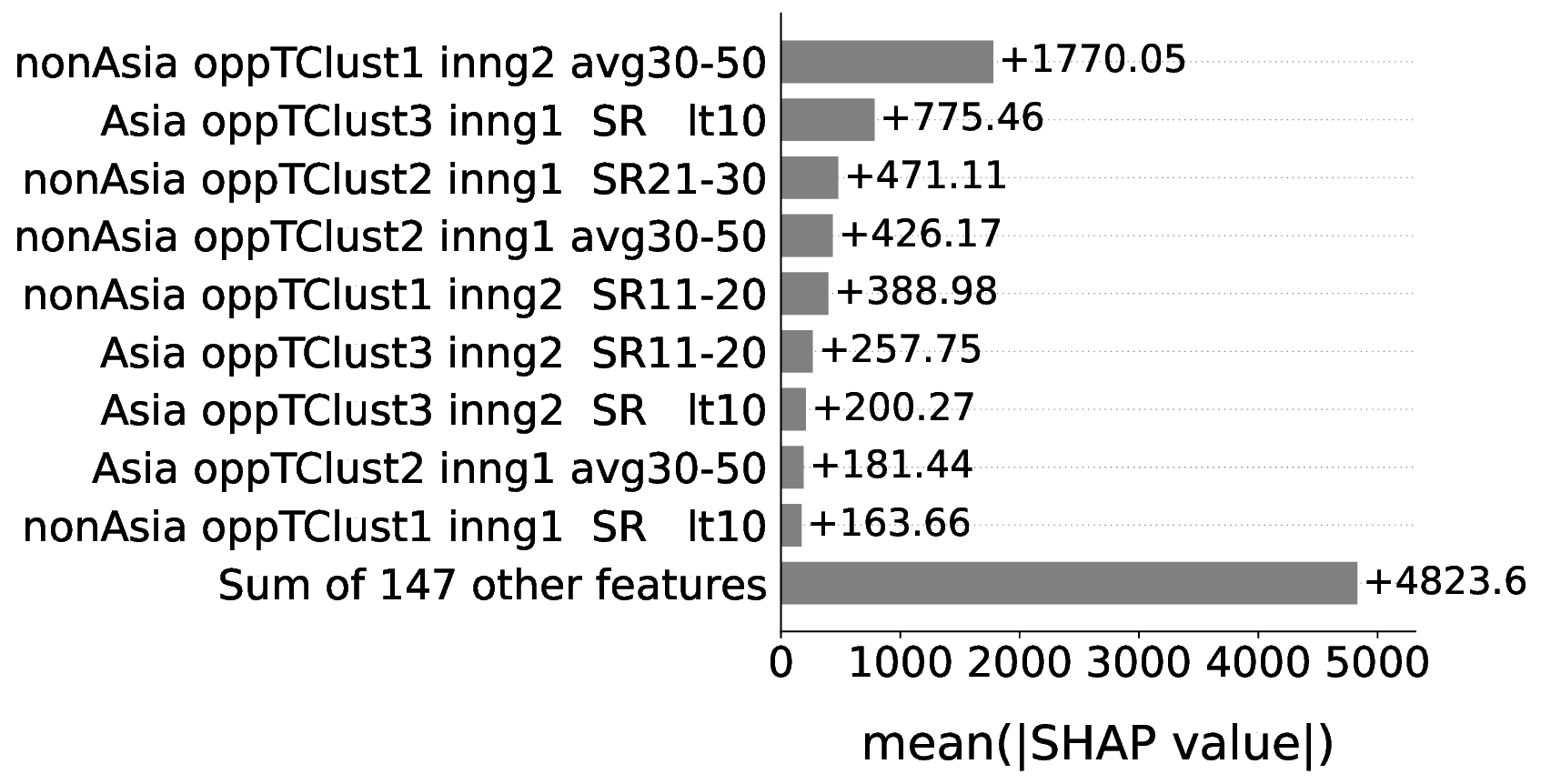}
   \caption{Bowlers}
   \label{fig_shap_bowlers}
\end{subfigure}
\caption{Mean absolute value of SHAP values for batters features \textbf{(a)} shows that runs scored against top batting teams at non-Asian venues is the most important feature. For bowlers \textbf{(b)} bowling strike rate in non-Asian venues is most significant.}
\label{fig_shap}
\end{figure}

We apply SHAP (SHapley Additive exPlanations) analysis~\citep{lundberg2017unified} to quantify the significance of features in determining the final prediction of the model. SHAP analysis runs a large number of predictions and compares the impacts of each feature. For SHAP analysis, we used bowlers and batters feature vectors against the aggregated quarterly ICC ratings over the last $19$ years. Figure~\ref{fig_shap_batsmen} shows that runs scored by the batter against top batting teams in Non-Asian venues are the most important feature for the batter. The Bowling strike rate in Non-Asian venues is the most important feature for the bowler, as shown in Figure~\ref{fig_shap_bowlers}. 

\subsection{Validation of Venue-wise Distribution of Matches}\label{Venue_Split_Validationa}
\begin{figure}[h!]
    \centering
    \begin{subfigure}{.44\textwidth}
        \centering
        \includegraphics[width=.95\linewidth]{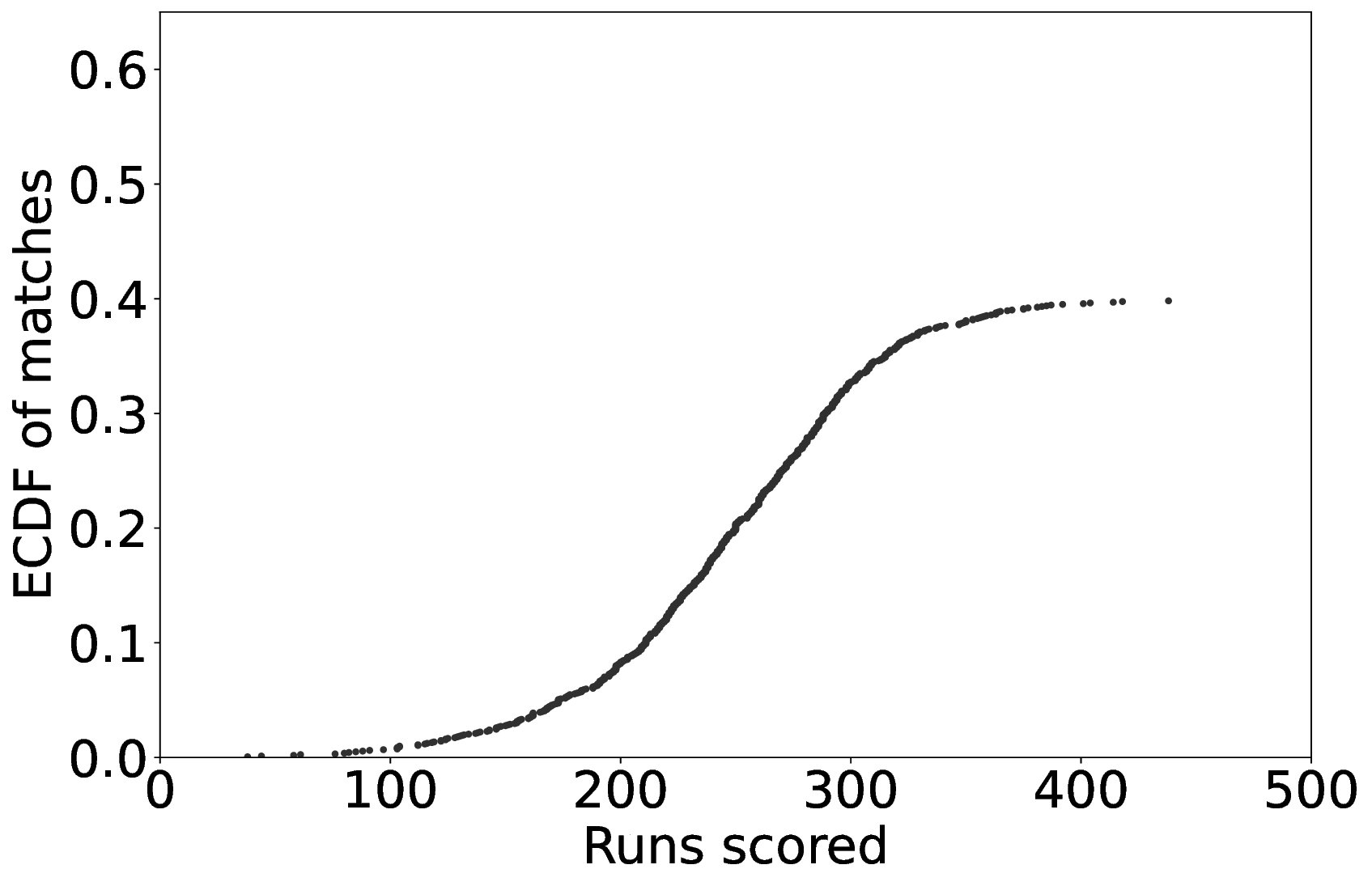}
        \caption{1st inning scores at Asian venues}
        \label{figure_innings_1_asia_ecdf}
    \end{subfigure} \qquad
    \begin{subfigure}{.44\textwidth}
      \centering
        \includegraphics[width=.95\linewidth]{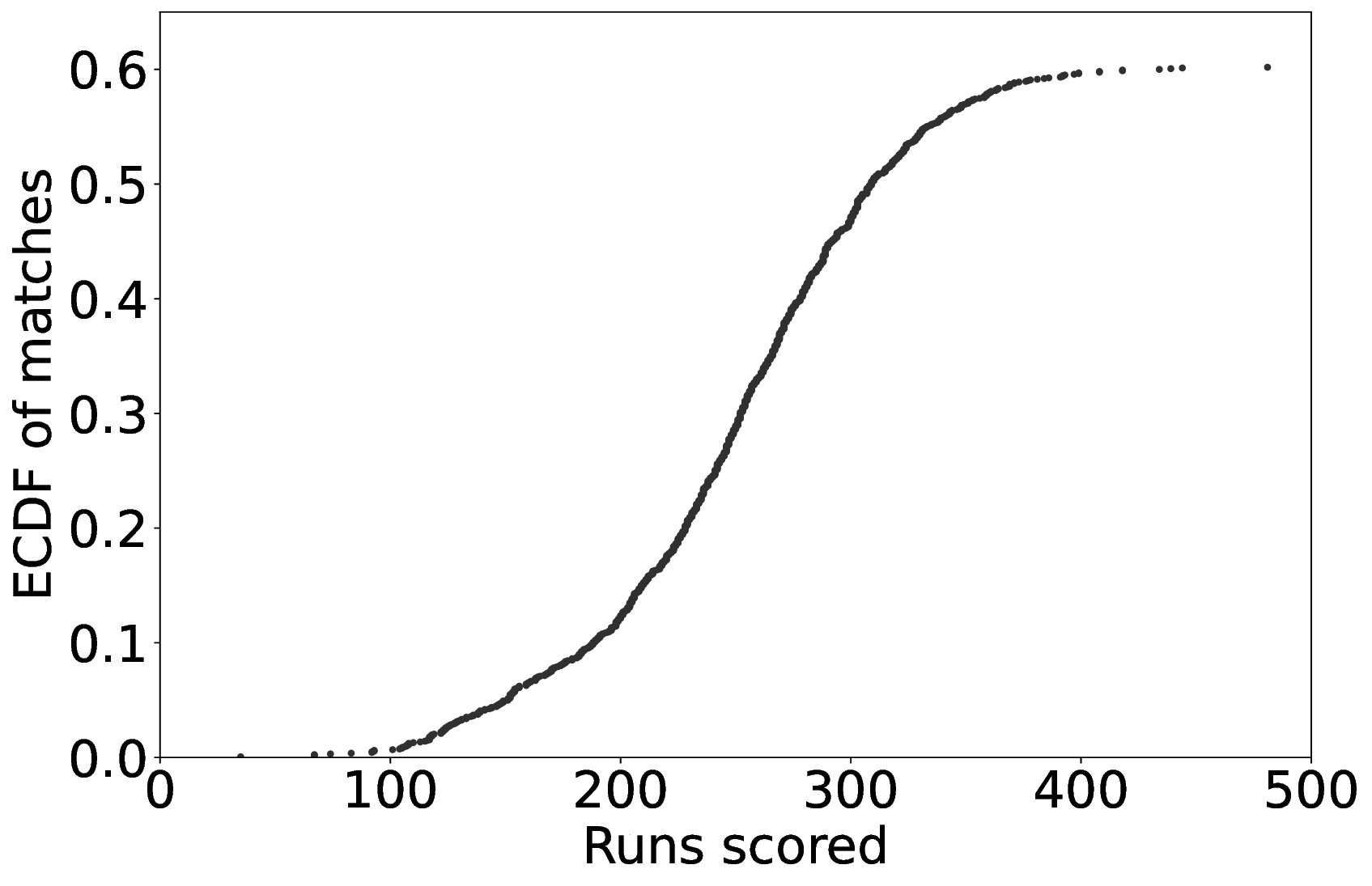}
        \caption{1st inning scores at Non-Asian venues}
        \label{figure_innings_1_nonasia_ecdf}
    \end{subfigure} \\
    \begin{subfigure}{.44\textwidth}
        \centering
        \includegraphics[width=.95\linewidth]{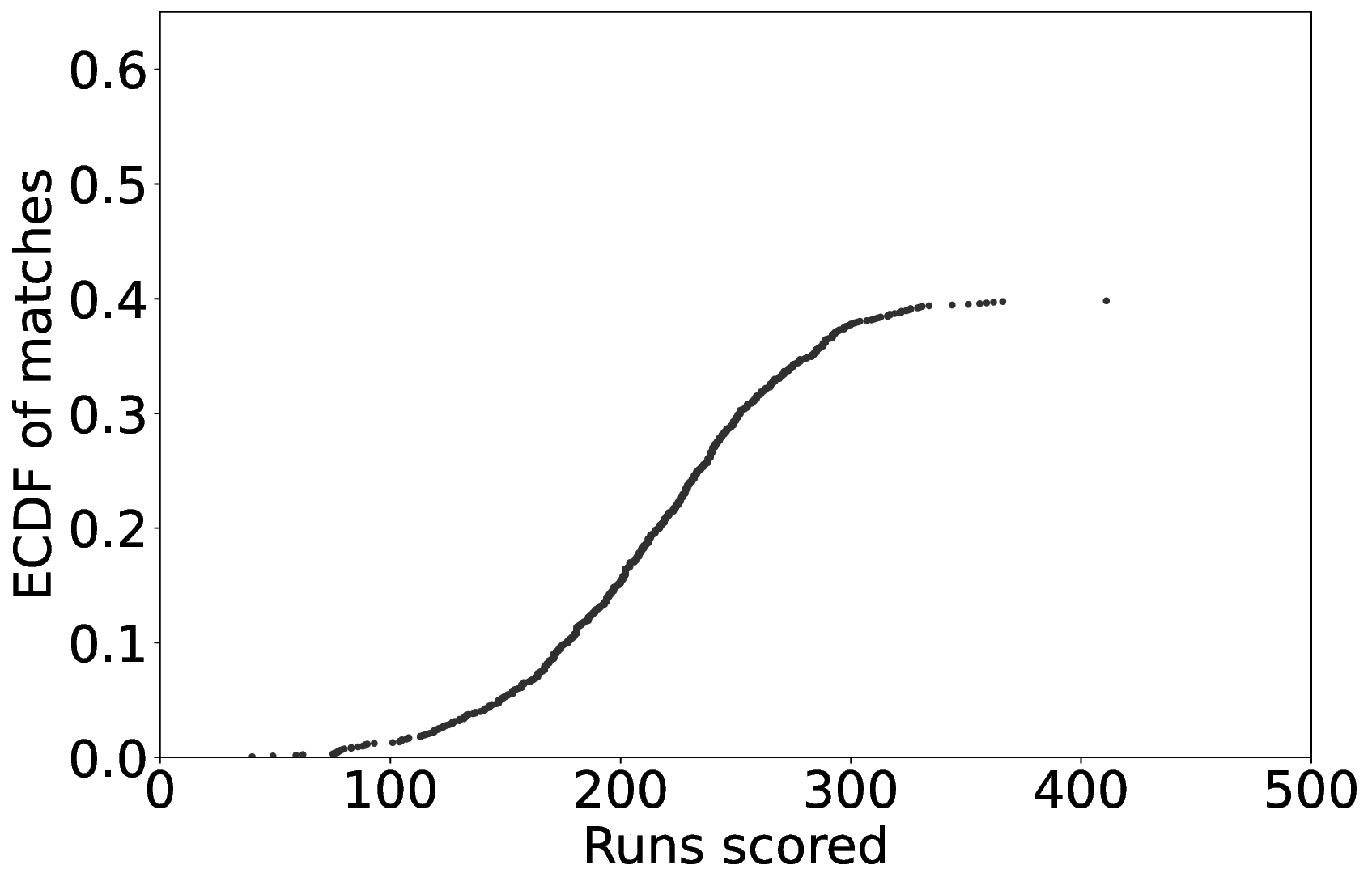}
        \caption{2nd inning scores at Asian venues}
        \label{figure_innings_2_asia_ecdf}
    \end{subfigure}\qquad
    \begin{subfigure}{.44\textwidth}
      \centering
        \includegraphics[width=.95\linewidth]{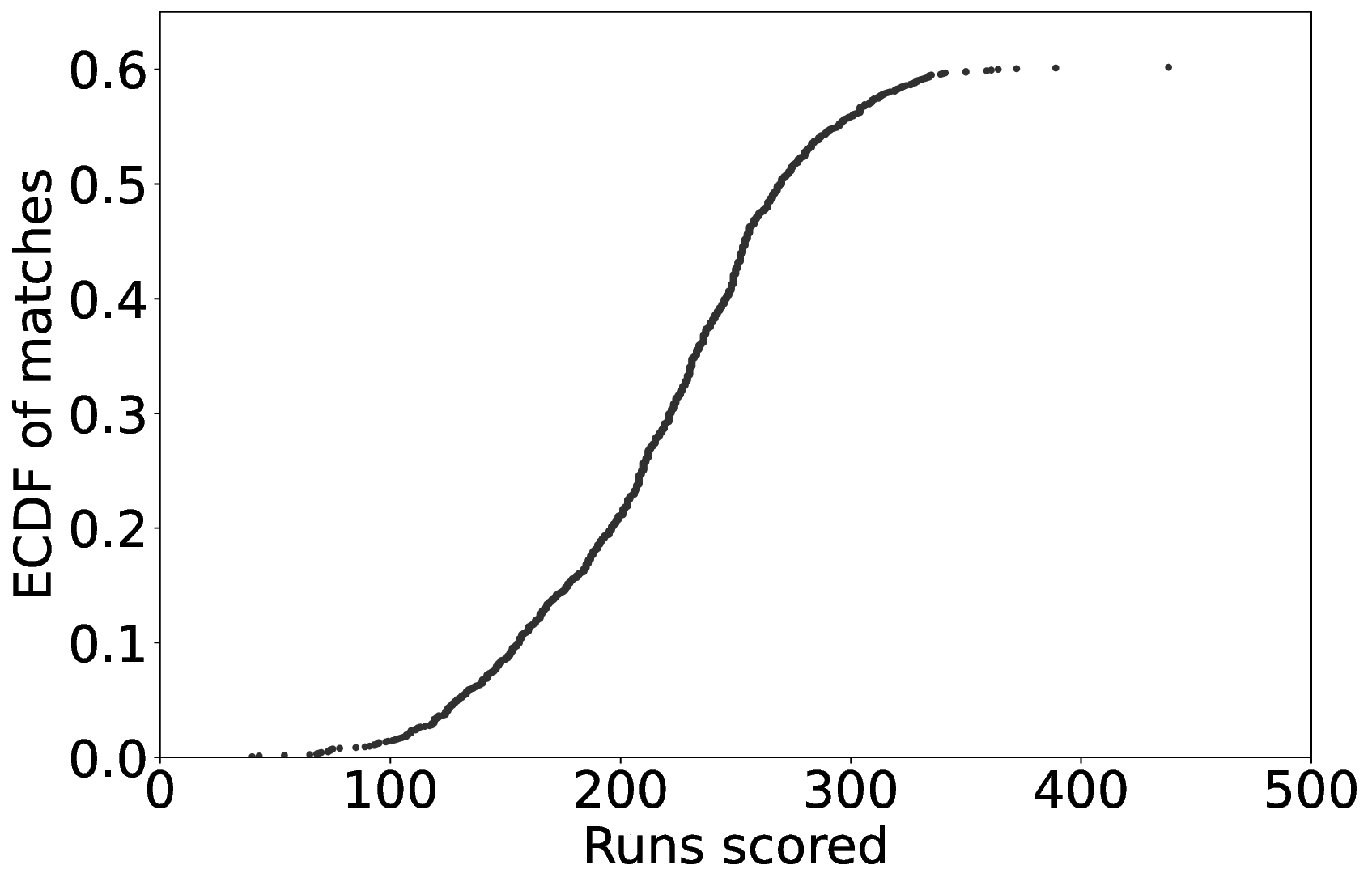}
        \caption{2nd inning scores at Non-Asian venues}
        \label{figure_innings_2_nonasia_ecdf}
    \end{subfigure}
    \caption{The ECDF plots showing the proportion of innings scores less than or equal to a given value for matches played on Asian and Non-Asian pitches. Innings scores on Asian pitches (\textbf{(a)} and \textbf{(c)}) exhibit substantially different patterns than those on Non-Asian pitches (\textbf{(b)} and \textbf{(d)}), as shown by the consistent gap between the curves.} 
    \label{fig_inningsRunsDistribution_ecdf}
\end{figure}

We demonstrate that scoring patterns vary significantly at different pitch conditions to validate the classification of match venues into Asian and non-Asian pitches. The empirical cumulative distribution function (ECDF) plots in Figure~\ref{fig_inningsRunsDistribution_ecdf} show the innings-wise cumulative distributions of scores in all matches on Asian and Non-Asian pitches. Significantly different distribution of total innings scores on Non-Asian and Asian venues justify distinguishing match venues for score projection.

\subsection{Evaluating Projected Remaining Runs} \label{projectedrunsevaluation}
\begin{figure}[h!]
	\centering
	\footnotesize
	\begin{tikzpicture}[scale=.95]
	\begin{axis}[
	ylabel shift=-1pt,
	xtick={1,5,10,15,20,25,30,35,40,45,50},
	ytick={0,5,10,15,20,25,30,35,40,45,50,60},
	height=0.4\columnwidth, 
	width=0.5\columnwidth, 
	ymin=0, ymax=50,
	legend columns=3,
	mark repeat={5},
	legend style={
		column sep=0.8ex,
	},	
	legend entries={\textsc{Our},\textsc{$Lewis_{235}$},\textsc{$Lewis_{MeanRuns}$},\textsc{$Lewis_{MedianRuns}$}},
	legend to name=commonlegend1,
	ylabel style={align=center},
	ylabel={Mean Absolute Error (runs) },
	xlabel={(a) Overs (First Innings)},
	]
	\addplot+[blue, mark size=2pt, mark options={blue},mark =triangle] table[x={over},y={MeanAbsError}, col sep=comma]{Figures/inprogressOfInning1_projectionError_allMatches.csv};

	\addplot+[black, mark size=2pt, mark=o, mark options={black}] table[x={over},y={Mean_absError_randForest}, col sep=comma]{Figures/inprogressOfInning1_projectionError_allMatches.csv};
	
	\addplot+[orange, mark size=2pt, mark=square, mark options={orange}] table[x={over},y={Mean_absError_ridgRegr}, col sep=comma]{Figures/inprogressOfInning1_projectionError_allMatches.csv};
	
	\addplot+[red, mark size=3pt, mark=asterisk, mark options={red}] table[x={over},y={MeanAbsErrorLewis35}, col sep=comma]{Figures/inprogressOfInning1_projectionError_allMatches.csv};
	\end{axis}
	\end{tikzpicture}
	\begin{tikzpicture}[scale=.95]
	\begin{axis}[
	ylabel shift=-1pt,
	xtick={1,5,10,15,20,25,30,35,40,45,50},
	ytick={0,5,10,15,20,25,30,35,40,45,50,60},
	height=0.4\columnwidth, 
	width=0.5\columnwidth, 
	ymin=0, ymax=50,
	legend columns=4,
	mark repeat={5},
	legend style={
		column sep=0.8ex,
	},	
	legend entries={\textsc{$k$NN},\textsc{RandomForest},\textsc{Ridge Regression},\textsc{lnc}},
	legend to name=commonlegend,
	ylabel style={align=center},
	ylabel={},
	xlabel={(b) Overs (Second Innings)},
	]
	\addplot+[blue, mark size=2pt, mark options={blue},mark = triangle] table[x={over},y={MeanAbsError}, col sep=comma]{Figures/inprogressOfInning2_projectionError_allMatches.csv};
	
	\addplot+[black, mark size=2pt, mark=o, mark options={black}] table[x={over},y={Mean_absError_randForest}, col sep=comma]{Figures/inprogressOfInning2_projectionError_allMatches.csv};
	
	\addplot+[orange, mark size=2pt, mark=square, mark options={orange}] table[x={over},y={Mean_absError_ridgRegr}, col sep=comma]{Figures/inprogressOfInning2_projectionError_allMatches.csv};
	
	\addplot+[red, mark size=3pt, mark=asterisk, mark options={red}] table[x={over},y={MeanAbsErrorLewis35}, col sep=comma]{Figures/inprogressOfInning2_projectionError_allMatches.csv};
	\end{axis}
	\end{tikzpicture}
	\\~\ref{commonlegend} 
	\caption{\textsc{mae} in projected remaining ($R(S_i)$) and actual $A(S_i)$ scores for both innings. $R(S_i)$ is predicted using $k$NN, Random Forest, Ridge Regression, and \textsc{lnc}.} 
	
	 \label{fig_inningsProjectionError}
\end{figure}
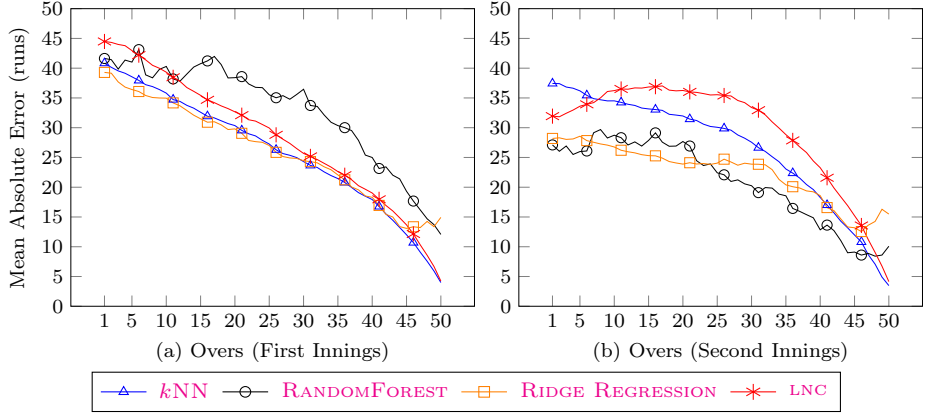

This section describes the accuracy of the computation of the projected scores by \textsc{camp}. We compute the mean absolute error (\textsc{mae}) in the projected scores $R(S_i)$ and the actual runs scored $A(S_i)$ by \textsc{camp} using $k$NN, Random Forest and Ridge Regression, and \textsc{lnc}. Figure~\ref{fig_inningsProjectionError} shows the \textsc{mae} in projected runs using \textsc{camp} (by applying $k$NN, Random Forest, and Ridge Regression) and using \textsc{lnc}. 

Figure~\ref{fig_inningsProjectionError}(a) shows that our $k$NN and Ridge Regression approaches outperform \textsc{lnc} throughout the first innings. However, the Random Forest is not as good as the inning proceeds. Figure~\ref{fig_inningsProjectionError}(b) shows the performance of our model and its comparison with \textsc{lnc} for the second innings. In the second innings, since \textsc{lnc} uses the same resource table (as the first innings), the error for \textsc{lnc} is higher. Since \textsc{camp} also considers the target remaining, it remains better at the start of the second innings (for $k$NN and Random Forest). For $k$NN, since we have the same resources at the beginning of the second innings, but the target is different, the error is higher as the feature vector does not have enough information. Also, the standard deviation of second innings runs is high, making it difficult for $k$NN to achieve higher accuracy at the start of the second innings. However, as the overs progress, the richer feature vectors for the $k$NN improve accuracy.

\vskip-.2in
\subsection{Evaluating Players' Ratings} \label{ratingevalution}
Evaluating the performance of \textsc{camp} is challenging as no objective ground truth exists for all players' contributions in a match. \textsc{lnc} gives some idea about the players' rankings, which is somewhat similar to ours, and \textsc{MoM} only identifies the {\em ``top-rated''} player. We evaluate \textsc{camp} in three aspects. 
\begin{enumerate}\setlength{\itemsep}{1pt}
    \item Firstly, we present a case study of a single match and show how our measure captures the context and quality of the opponent batter or bowler as opposed to the standard performance measure. 
    \item We then report the agreement of our top contributor with \textsc{MoM} and compare this agreement with that of \textsc{lnc}. 
    \item Finally, we compare the performance of \textsc{camp} on the case study of a series reported by \textsc{lnc}.
\end{enumerate}

\subsubsection{Comparison with Traditional Batting and Bowling Performance Measures}

Traditional performance measures of batting and bowling offer no objective way to incorporate the situation in which runs are scored or conceded. For example, two batters scoring the same number of runs in the same number of deliveries at different stages of games facing different types of bowlers are not valued equally, and nearly always, some verbal qualification is required to place the statistics into context. We show how \textsc{camp} caters to this limitation through the case study of a randomly selected match between NZ and PAK on October 25, $2006$ at Mohali\footnote{\href{https://www.espncricinfo.com/series/icc-champions-trophy-2006-07-232694/new-zealand-vs-pakistan-14th-match-249752/full-scorecard}{Full Scorecard of NZ vs. PAK 14th Match in ICC Champions Trophy (2006/07) - https://www.espncricinfo.com/series/232694/scorecard/249752/}}.

In this game, Fleming scored $80$ runs (strike rate $76.10$) and was declared \textsc{MoM}, which is not obvious from the scorecard (Table~\ref{tab:NZvsPakBattingInnings} for the scorecards). Styris scored the highest runs ($86$ (strike rate $76.19$)) with the highest number of boundaries in his batting. Bond took the highest wickets ($3$)(economy $4.50$). Oram scored $31$ runs (strike rate  $119.23$), which is more than the strike rate of Styris and Fleming. Also, Oram took $2$ wickets with the highest economy ($3.12$). The top performer (Fleming) is not obvious from the scorecard only. However, the context-aware \textsc{camp} offers more meaningful insights (Table~\ref{tab:NZvsPakContributions}). Fleming (\textsc{MoM}) has the highest \textsc{camp$_{score}$}, which agrees with experts' decision of \textsc{MoM}. In this case study, \textsc{camp} also outperforms \textsc{lnc}. According to \textsc{lnc}, Oram is the best contributor, and Fleming (\textsc{MoM}) is ranked $2nd$ in the winning team ($3rd$ among all $22$ players). Also, note that Styris and Bond are declared the best-performing batter and bowler by ESPNcricinfo, respectively.

\begin{table}[h!]
	\centering
    \begin{minipage}[t]{0.485\textwidth}
    \vspace{0pt}
    \resizebox {1\textwidth} {!} {%
    	\begin{tabular}{lcccccl}
    	\toprule
    	Player & Team & Runs  & Balls & ${4s}$ & ${6s}$ & Out by \\%
       \toprule
        \textbf{S. Fleming} & NZ & 80   & 105   & 8  & 1 & S. Malik   \\%
        P. Fulton & NZ     & 7    & 14    & 1  & 0  & I. Anjum \\%
        S. Styris & NZ     & 86   & 113   & 10 & 0 & I. Anjum  \\%
        J. Oram  & NZ      & 31   & 26    & 4  & 1  & U. Gul       \\%
        B. McCullum & NZ & 27   & 13    & 3  & 1  & S. Malik   \\%
        J. Franklin & NZ   & 9    & 5     & 1  & 0    & not out     \\%
        M. Yousuf      & PAK  & 71   & 92    & 9  & 0  & S.  Fleming      \\
        \bottomrule
        \end{tabular}%
    }
    \end{minipage}
    \quad
    \begin{minipage}[t]{0.48\textwidth}
    \vspace{0pt}
    \resizebox {1\textwidth} {!} {%
    	\begin{tabular}{lccccc}
    	\toprule
    	Player & Team & Overs & Runs & Wickets & Economy \\
        \toprule
        K. Mills     & NZ & 7.3 & 38 & 2 & 5.06 \\
        S. Bond      & NZ & 10  & 45 & 3 & 4.50 \\
        J. Franklin  & NZ & 9   & 47 & 1 & 5.22 \\
        J. Oram      & NZ & 8   & 25 & 2 & 3.12 \\
        D. Vettori  & NZ & 10  & 52 & 1 & 5.20 \\
        N. Astle    & NZ & 2   & 11 & 0 & 5.50 \\
        S. Malik    & PAK & 5  & 25 & 1 & 5.00 \\
    	\bottomrule
        \end{tabular}%
    }
    \end{minipage}
    \caption{Batting and bowling scorecards of the randomly selected NZ vs. PAK (2006) match due to non-obvious \textsc{MoM} (S. Fleming) from the winning team's (NZ) scorecards.} %
    \label{tab:NZvsPakBattingInnings}
\end{table}

\begin{table}[h!]
    \scriptsize
    \centering
	\begin{tabular}{lccccc}
	\toprule
	Player & Team & \textsc{camp$_{score}$} & \textsc{camp$_{rank}$} & \textsc{lnc$_{score}$} & \textsc{lnc$_{rank}$}\\
    \toprule
    \textbf{S. Fleming}  & NZ & $+35.4$ & 1 & $+28.77$  & 3\\ %
    S. Bond  & NZ & $+15.4$  &  2 & $+28.26$  & 4\\ %
    J. Oram   & NZ & $+11.2$ &  4 & $+36.55$  & 1\\ %
    S. Styris  & NZ & $+10.5$ & 5 & $+13.62$  & 7\\ %
    B. McCullum  & NZ & $+6.2$ & 7 & $+11.02$  & 8\\ %
    K. Mills  & NZ & $+0.56$ &   10 & $-7.22$ & 12\\ %
    M. Yousuf & PAK & $+12.7$ & 3 & $+34.81$   & 2\\ %
    M. Hafeez & PAK  & $+10.0$ &  6 & $+7.515$ & 9\\ %
    S. Malik  & PAK & $+5.82$ &  8 & $+19.83$  & 5\\ %
    K. Akmal  & PAK & $+5.0$  &  9 & $+14.36$ & 6\\ %
	\bottomrule
    \end{tabular}
    \caption{\textsc{camp} ratings of prominent performers from both teams in the randomly selected SA vs. IND (2001) match due to non-obvious \textsc{MoM} (S. Fleming).}
    \label{tab:NZvsPakContributions}
\end{table}

We also show that if the top contributor by \textsc{camp} disagrees with \textsc{MoM}, the difference between \textsc{camp} ratings among top-rated players is very small. A case study of a randomly selected match between SA and IND on October 26, $2001$ at Durban is used to evaluate the contribution difference between top players for non-obvious \textsc{MoM}\footnote{ \href{https://www.espncricinfo.com/series/8660/scorecard/66107/south-africa-vs-india-final-standard-bank-triangular-tournament-2001-02}{Scorecard: SA vs. IND {\tiny SB Triangular Tournament(01/02)-https://www.espncricinfo.com/series/8660/scorecard/66107/}}}.

In this game, the \textsc{MoM} (S. Pollock from the winning team) is not obvious from the scorecard (Table~\ref{tab:SAwithIndiaBattingInnings}). Kirsten scored $87$ runs in $108$ balls, Kemp took $3$ wickets with economy $3.15$ and Pollock took $2$ wickets with economy $2.11$. The top performer is not obvious from the scorecard. However, the context-aware \textsc{camp} offers more meaningful insights (Table~\ref{tab:SAwithIndiaContributions}). Kirsten has the highest \textsc{camp$_{score}$}, followed by Kemp and Pollock with a very slight difference. However, Pollock was awarded \textsc{MoM}. It is important to note that the contribution difference between Pollock and the players above him is very little. If \textsc{MoM} is not the top contributor, this may be due to experts' subjective judgment that considers other factors such as fielding, captaincy, and wicket-keeping.

\begin{table}[h!]
    \centering
    \begin{minipage}[t]{0.475\textwidth}
    \vspace{0pt}
    \resizebox {1\textwidth} {!} {%
    	\begin{tabular}{lcccccl}
    	\toprule
    	Player & Team & Runs  & Balls & ${4s}$ & ${6s}$ & Out by \\
       \toprule
    	G. Kirsten  & SA & $87$ & $108$ & $9$ & $1$ &   	H. Singh \\
    	J. Kallis   & SA & $39$ & $63$ & $5$ & $0$ &   S. Tendulkar  \\
        \textbf{S. Pollock}  & SA & $0$ & $4$ & $0$ & $0$ &   Not Out\\	
        S. Ganguly & IND & $9$ & $17$ & $1$ & $0$ &    S. Pollock \\
    	R. Dravid & IND & $77$ & $102$ & $6$ & $0$ &    J. Kemp\\
    	Y. Singh & IND & $2$ & $3$ & $0$ & $0$ &   J. Kemp   \\
    	A. Kumble  & IND & $0$ & $2$ & $0$ & $0$ &   J. Kemp  \\
        \bottomrule
        \end{tabular}%
    }
    \end{minipage}
    \quad
    \begin{minipage}[t]{0.47\textwidth}
    \vspace{0pt}
    \resizebox {1\textwidth} {!} {%
    	\begin{tabular}{lccccc}
    	 \toprule
    	Player & Team & Overs & Runs & Wickets & Economy \\
        \toprule
    	\textbf{S. Pollock} & SA & $9$ & $19$ & $2$ & $2.11$ \\
    	J. Kemp & SA& $6.2$ & $20$ & $3$ & $3.15$  \\ 	
    	N. Hayward  & SA & $10$ & $38$ & $2$ & $3.80$ \\ 	
    	J. Kallis   & SA & $8$ & $41$ & $0$ & $5.12$ \\ 	
    	L. Klusener & SA & $5$ & $19$ & $1$ & $3.80$  \\ 
    	H. Singh    & IND & $10$ &	$48$ & $2$  & $4.80$ \\
        S. Tendulkar & IND & $5$ &  $27$ & $2$  & $5.40$  \\
    	\bottomrule
        \end{tabular}%
    }
    \end{minipage}
    \caption{Batting and bowling scorecards of the randomly selected SA vs. IND (2001) match due to non-obvious \textsc{MoM} (S. Pollock) from the winning team's (SA) scorecards.}
    \label{tab:SAwithIndiaBattingInnings}
\end{table}

\begin{table}[h!]
    \scriptsize
    \centering
	\begin{tabular}{lccccccc}
	\toprule
	Player & Team & ${C_{bat}}$ & ${C_{bowl}}$   & {\textsc{camp$_{score}$}}& {\textsc{camp$_{rank}$}} & {\textsc{lnc$_{score}$}} & {\textsc{lnc$_{rank}$}} \\
    \toprule
	G. Kirsten  & SA & $+12.95$ & $0$ & $+12.95$ & $2$ & $+23.32$ & $1$ \\
	
	J. Kemp  & SA & $0$ &  $+64.20$ & $+12.84$ & $3$	& $+19.72$ & $3$\\
	
	\textbf{S. Pollock} & SA &  $+0.10$ & $+61.10$ & $+12.22$ & $4$ & $+22.83$ & $2$\\
	
	N. Hayward & SA & $0$ &$+55.00$ & $+11.00$ & $6$ & $+11.35$ & $5$\\	
	L. Klusener & SA & $0$  &  $+20.30$ & $+4.06$ & $12$ & $+5.62$ & $7$\\
	
	J. Kallis & SA & $-14.08$ & $-17.00$ & $-17.48$ & $16$  & $-24.33$ & $22$ \\ 
	
	R. Dravid & IND & $+17.02$ & $0$ & $+17.02$ & $1$ & $+16.61$ & $4$ \\
	\bottomrule
    \end{tabular}
    \caption{\textsc{camp} ratings of prominent performers from both teams in the randomly selected SA vs. IND (2001) match due to non-obvious \textsc{MoM} (S. Pollock).}
    \label{tab:SAwithIndiaContributions}
\end{table}

\subsubsection{Comparison with Man of the Match (\textsc{MoM})}
The man of the match (\textsc{MoM}) is nominated through a rigorous subjective process by field experts who observe the match closely. The highest net contributor by \textsc{camp} closely agrees with the \textsc{MoM}. We report the {\it agreement accuracy} (fraction of matches where the top contributor by \textsc{camp} is the \textsc{MoM}). We implemented \textsc{lnc} technique to select the top contributor\footnote{\url{https://github.com/sohaibayub/CAMP}}.
Table~\ref{tab:manOfMatchamong11} shows that \textsc{camp} outperforms \textsc{lnc} in agreement accuracy. The agreement accuracy of \textsc{camp} is $66\%$ , $83\%$ and $90\%$ for \textsc{MoM} having rank 1, rank 2 and rank 3 on \textsc{camp} scale, respectively. To the best of our knowledge, this is the highest \textsc{MoM} agreement accuracy reported for \textsc{odi} cricket.

\begin{table}[h!]
\centering
\resizebox{\textwidth}{!}{%
	\begin{tabular}{@{\extracolsep{4pt}}lccccc@{}}
		\toprule
	 	 & \multicolumn{2}{c}{{$11$ players of winning team}}	& \multicolumn{2}{c}{{$22$ players of both teams}} \\
		 \cmidrule{2-3} \cmidrule{4-5}
		    &  \textsc{camp} & \textsc{lnc}   &  \textsc{camp} & \textsc{lnc} \\
		  \midrule
	\textsc{MoM} having rank $1$ on \textsc{camp} scale & $638(66.3\%)$ & $585 (60.8\%)$ & $458(47.6\%)$ & $461(47.9\%)$\\
	\textsc{MoM} among top $2$ on \textsc{camp}  scale & $799(83.1\%)$ & $784(81.5\%)$  & $686(71.3\%)$ & $650(67.6\%)$\\
	\textsc{MoM} among top $3$ on \textsc{camp} scale & $867(90.2\%)$ & $864(89.9\%)$ & $789(82.1\%)$ & $773(80.4\%)$ \\
  	\bottomrule
	\end{tabular}
	}
	\captionsetup{font={small}}
	\caption{Comparison with \textsc{MoM} in $961$ matches among the $11$ winner team and all $22$ players.}
	\label{tab:manOfMatchamong11}
\end{table}

It is well known that \textsc{MoM} is mostly from the winning team. Therefore, we report results for \textsc{MoM} rank among the winning team players and all $22$ players of both teams separately in Table~\ref{tab:manOfMatchamong11}. As the accuracy of \textsc{MoM} being the top contributor among $22$ players is relatively low, we have observed that out of total $961$ matches, there are $228$ such matches, where \textsc{MoM} is ranked second among all $22$ players. However, out of these $228$ matches, \textsc{MoM} is the top contributor of his team in $154$ matches, which shows the bias toward selecting \textsc{MoM} from the winning team. 

\subsubsection{Comparison with \textsc{lnc} on Series Level}
Similar to \textsc{MoM}, ICC also announces \textbf{P}layer \textbf{o}f the \textbf{S}eries (\textsc{PoS}) based on the overall performance of  participating players through the series (tournament). \textsc{camp} evaluates players' contributions in each match of a series. Since there is no other baseline metric to validate the ratings of players at a series level, we utilize the accuracy of the agreement between the (aggregated) top contributor of the series and \textsc{PoS}.~\citep{Lewis05} evaluated \textsc{lnc} on the \textbf{V}ictoria \textbf{B}itter VB Series (2002-03) played between ENG, AUS and SL.
The contribution scores aggregated over the $14$ matches by \textsc{camp} and by \textsc{lnc} are given in Table~\ref{tab:VBSeriescomparisonwithLewis}.

\begin{table}[h!]
    \small
	\centering
	\begin{tabular}{lccccc}
        \toprule
		{Player} & {Team} & {\textsc{camp$_{score}$}} & {\textsc{camp$_{rank}$}} & {\textsc{lnc$_{score}$}} & {\textsc{lnc$_{rank}$}} \\
		\midrule
		 S. Jayasuriya & SL & $89.86$  & $1$  & $97.18$  & $4$  \\
		 P. Collingwood  & ENG & $65.66$  & $2$ & $110.94$  & $2$   \\
		 \textbf{B. Lee} & AUS & $65.42$  & $3$  & $33.99$  & $14$   \\
		 A. Bichel & AUS & $50.89$  & $4$  & $45.90$  & $10$  \\
		 B. Williams & AUS & $49.50$  & $5$ & $29.77$  & $15$   \\
		 D. Lehmann  & AUS & $48.32$  & $6$ & $75.62$  & $5$  \\
		 A. Gilchrist  & AUS & $46.63$  & $7$ & $105.25$  & $3$   \\
		 M. Hayden & AUS & $35.33$  & $8$ & $152.76$  & $1$   \\
		 A. Caddick & ENG & $32.82$  & $9$ & $56.00$  & $6$   \\
		 N. Bracken  & AUS & $31.00$  & $10$ & $48.29$  & $8$   \\
		\bottomrule
	\end{tabular}
	\caption{Comparison of scores and ranks by \textsc{camp} and \textsc{lnc} for top $10$ players in VB series (02-03). \textsc{lnc$_{score}$} are reported in~\citep{Lewis05}. Brett Lee was the \textsc{PoS}.}
	\label{tab:VBSeriescomparisonwithLewis}
\end{table}

In this series, \textsc{PoS} nominated by ICC (Brett Lee)\footnote{\href{https://www.espncricinfo.com/series/vb-series-2002-03-61122/australia-vs-england-2nd-final-65642/full-scorecard}{Player of the Series announced by ICC}} is the top $1$ for the series-winning team (AUS) and among the top $3$ for all matches by \textsc{camp}. However, \textsc{lnc} places him at the $14^{th}$ position. This analysis exhibits that \textsc{camp} is more effective than \textsc{lnc} for players' contributions at the series level as well. Players' contributions at the tournament level can also be computed using the same method.
The overall \textsc{MoM} agreement of our proposed model (for the VB-series) is given in Table~\ref{tab:manOfMatchVBseries}.

\begin{table}[h!]
	\small
	\centering
	\begin{tabular}{lcc}
	\toprule
      &  {Agreement} & {Agreement Accuracy}  \\  
    \toprule
	\textsc{MoM} ranked $1^{st}$ by \textsc{camp} & $10$ times & \textbf{71.14 \%}\\
	\textsc{MoM} ranked among top $2$ by \textsc{camp} & $12$ times & \textbf{85.71 \%} \\
    \bottomrule
		\end{tabular}
\caption{\textsc{camp} rankings of \textsc{MoM} for the $14$ matches in VB series (2002-2003).} %
	\label{tab:manOfMatchVBseries}
\end{table}

\section{Conclusion}\label{conclusion}
We proposed the \textsc{camp} measure to objectively quantify players' performance and assess players'  contribution to a cricket game. \textsc{camp}'s data-driven players rating achieves close agreement with the man of the match awards. Our approach can be extended to any format of cricket. An individual player's contribution is measured based on the game's context and the opposition's strength. Each stage of the innings demands a different nature of play, and expectations from players and their performances change over time. Our framework keeps track of the current match situation and assigns context-aware ratings to the players. 
In the future, \textsc{camp} can be extended  to incorporate additional factors such as (mis)fielding (including catches, run-outs, and stumpings), extras (i.e., Bye and Leg bye), running to distinguish wicket loss, changes in rules during the power play, captaincy, and wicket-keeping. This can be accomplished by utilizing text analytics techniques on match commentary and social media feedback provided by the spectators. 
Another potential area of research is to leverage video analytics on highlights, speech processing to analyze the crowd's cheering, and sentiment analysis on social media feeds related to matches. This out-of-ground data can provide valuable insights into evaluating players' contributions to match outcomes. It would also be interesting to explore the potential usefulness of \textsc{camp}'s prediction subroutine for betting purposes and evaluate the performance of our prediction subroutine using betting-related metrics, such as profitability or return on investment.

\pagebreak

\appendix
\section{Rules and Objectives of One Day International Cricket Game}\label{appendix_section}
This section presents an overview of the objective and basic rules of the \textsc{odi} cricket game, along with a few basic terminologies.

\paragraph*{Toss:}
\vskip-.15in
As in other sports, a cricket match starts with a toss. The toss-winning team has the choice to bat first or ask the opponent to bat first. This important decision is made considering the nature of the field, weather conditions, and the teams' relative strengths.
\paragraph*{Objective:}\vskip-.15in

A match is played between two teams of $11$ players each. Suppose $team_A$ is batting first, at the start of the first innings, $team_A$ has $50$ overs and $10$ wickets to score the maximum runs before either $50$ overs are completed or $10$ wickets are lost. An over consists of $6$ balls to be bowled by any player of the second team, $team_B$. The other $10$ players spread in the field to stop as many runs as possible. A bowler can bowl a maximum of $10$ overs in an innings. Runs are scored by hitting the ball and exchanging positions between two batters or hitting the ball outside the boundary for $4$ and $6$. $Team_B$ starts its innings with the same resources (overs and wickets). However, $team_B$ has to chase the target ($team_A$'s score plus one) to win. The second innings finishes when the resources are consumed or the target is achieved, whichever happens first.

\paragraph*{Wicket Loss:}\vskip-.15in

A batter can lose his wicket in several pre-defined ways, such as bowled, caught by opponents, run-out, or Leg Before Wicket (LBW).

\paragraph*{Target Runs:}\vskip-.15in

The number of runs accumulated by $team_A$ after the first innings plus $1$ is set as a target for the $team_B$ batting in the second innings.

\paragraph*{Match Outcome:}\vskip-.15in
The team with the highest score is declared the winner if both innings are completed without interruption (rain or other severe weather conditions).

\paragraph*{Resources:} \vskip-.15in

A team batting first has $10$ wickets and $50$ overs collectively called resources. $Team_A$ tries to maximize runs while consuming the resources. The first innings comes to an end when either of the resources finishes.

\paragraph*{Duckworth-Lewis Resource Table:}\vskip-.15in

The \textsc{dl} resource table (Table~\ref{tab:DLS_Table}) represents the mean percentage of further runs scored with $w$ wickets lost and $u$ overs left. For an average \textsc{odi}, the total score of team $1$ is $235$. Readers are referred to~\citep{Duckworth1998,duckworth2004,Lewis05} (and the references therein) for details.

\begin{table}[h!]
\centering
\footnotesize
	\begin{tabular}{@{\extracolsep{4pt}}ccccc@{}} \toprule
	\multirow{2}{*}{Overs left} & \multicolumn{4}{c}{Wickets lost} \\
		\cmidrule{2-5}
		  &  $0$ & $2$  & $4$ & $9$  \\ \midrule
        50 & $100$ & $83.8$ & $62.4$ & $7.6$\\
        40 & $90.3$ & $77.6$ & $59.8$ & $7.6$\\
        30 & $77.1$ & $68.2$ & $54.9$ & $7.6$\\
        20 & $58.9$ & $54.0$ & $46.1$ & $7.6$\\
        10 & $34.1$ & $32.5$ & $29.8$ & $7.6$\\ \bottomrule
	\end{tabular}
	\caption{\textsc{dl} resource table showing the percentage of remaining expected scores with the number of overs left and wickets lost.}
	\label{tab:DLS_Table}
\end{table}

\end{document}